\documentclass[preprint, review,12pt]{elsarticle}

\usepackage{amssymb}
\usepackage{amsthm}
\usepackage{amsmath}
\usepackage{url}
\usepackage{subcaption}

\usepackage{algorithm}
\usepackage{algpseudocode} % it includes package algorithmicx
\usepackage{multirow}

\journal{Pattern Recognition}

\def\mat#1{\mathchoice{\mbox{\boldmath $\displaystyle\tt#1$}}
{\mbox{\boldmath$\textstyle\tt#1$}}
{\mbox{\boldmath$\scriptstyle\tt#1$}}
{\mbox{\boldmath$\scriptscriptstyle\tt#1$}}}

\def\vect#1{\mathchoice{\mbox{\boldmath $\displaystyle\bf#1$}}
{\mbox{\boldmath  $\textstyle\bf#1$}}
{\mbox{\boldmath  $\scriptstyle\bf#1$}}
{\mbox{\boldmath  $\scriptscriptstyle\bf#1$}}}

\def\v0{{\vect 0}}

\def\vn{{\vect n}}

\def\vv{{\vect v}}

\def\vx{{\vect x}}
\def\vy{{\vect y}}

\def\v0{{\vect 0}}

\def\mDelta{{\mat\mDelta}}

\def\mA{{\mat A}}
\def\mB{{\mat B}}

\def\mM{{\mat M}}

\def\m1{{\mat 1}}

\begin{document}

\begin{frontmatter}

\title{ELSED: Enhanced Line SEgment Drawing}

\author[lbl:graffter,lbl:upm]{Iago Su\'arez}
\author[lbl:urjc]{Jos\'e M. Buenaposada}
\author[lbl:upm]{Luis Baumela}
\affiliation[lbl:graffter]{organization={The Graffter},
            addressline={Campus Montegancedo s/n. Centro de Empresas},
            city={Pozuelo de Alarc{\'o}n},
            postcode={28223},
            country={Spain}}

\affiliation[lbl:upm]{organization={Departamento de Inteligencia Artificial. Universidad Polit{\'e}cnica  de Madrid},
            addressline={Campus Montegancedo s/n},
            city={Boadilla del Monte},
            postcode={28660},
            country={Spain}}
\affiliation[lbl:urjc]{organization={ETSII. Universidad Rey Juan Carlos},
            addressline={C/ Tulip{\'a}n, s/n},
            city={M{\'o}stoles},
            postcode={28933},
            country={Spain}}

\begin{abstract}
Detecting local features, such as corners, segments or blobs, is the first step in the pipeline of many Computer Vision applications. Its speed is crucial for real-time applications. 
In this paper we present ELSED, the fastest line segment detector in the literature. The key for its efficiency is a local segment growing algorithm that connects gradient-aligned pixels in presence of small discontinuities. 
%The  key  for  efficiency  is  a  new  method  that  incremen-tally connects pixels with strong gradients whereas it enforces them to lie instraight lines, saving time and improving results. 
%
The proposed algorithm not only runs in devices with very low end hardware, but may also be parametrized to foster the detection of short or longer segments, depending on the task at hand. 
% These improvements, make it amenable for low computational power devices such as smartphones and drones. 
%
%Furthermore ELSED allows to define the type of segments to be detected (i.e. short line segments or long ones) depending on the task at hand. 
%
We also introduce new metrics to evaluate the accuracy and repeatability of segment detectors. In our experiments with different public benchmarks we prove that our method accounts the highest repeatability and it is the most efficient in the literature\footnote{Source code: \url{https://github.com/iago-suarez/ELSED}}. In the experiments we quantify the accuracy traded for such gain.
\end{abstract}


\end{frontmatter}


\section{Introduction}
\label{sec:introduction}

Detecting segments and full lines in digital images is a recurrent problem in Computer Vision (CV). 
%these features arise from pixels that are located mostly on boundaries of geometric man-made objects such as buildings, roads, urban furniture, etc. 
% Motivation
Line segments play an important role in understanding the geometric content of a scene as they are a compressed and meaningful representation of the image content. Moreover, segments are still present in low-textured settings where the classical methods based on corners or blobs usually fail.
% Cases of usage
Segment detection has been employed in a large number of CV tasks such as
3D reconstruction~\cite{zhou2019learning, miraldo2018minimal}, %li2017line, hofer2017efficient, zeng2020bundle, ramalingam2013lifting
%~\cite{cheng2001three}~\cite{tang2006projective}
SLAM~\cite{li2020structure, gomez2019pl}, % pumarola2017pl
Visual Odometry~\cite{gomez2016pl}, % koletschka2014mevo, lu2015robust, camposeco2015using
3D camera orientation via Vanishing Points Detection~\cite{lezama2014finding, suarez2018fsg}, % bazin2012globally, lee2015real, zhou2017detecting
% crack detection in materials~\cite{prasanna2014automated}, 
cable detection in air-crafts~\cite{tian2021noncontact}, % yetgin2015comparison 
% biomedical images~\cite{santosh2017line} %, santosh2015stitched
or road detection in Synthetic Aperture Radar images~\cite{Liu2020LSDSAR}.

% Briefly speak about the history of Line and segment detection in the last years
Nowadays CV algorithms are ubiquitous and they are expected to run on resource-limited devices~\cite{suarez2021bad}. To this end, low-level algorithms such as the local feature detectors must be very efficient. %For example the FAST corner detector~\cite{rosten2006fast} takes a few milliseconds on a smartphone). 
Traditional global line detection approaches based on the Hough transform lack efficiency. Thus, various local methods emerged addressing the issue of efficiency. LSD~\cite{grompone2010lsd} was one of the first approaches to achieve excellent results with a local approach. 
%However, LSD is not fast enough for limited devices and to improve efficiency the 
Edge drawing methods further improve the efficiency~\cite{topal2012edge, akinlar2011edlines, zhang2021ag3line}. In a first step, they work by connecting edge pixels following the direction perpendicular to the gradient. In a second step, they fit the desired curve, a line in the simplest case, to these edges.% found in the first step. % following the image edges in the direction perpendicular to the gradient. 

The method presented in this paper improves on the drawing methods by fitting a line segment to the connected edge pixels and using its direction to guide the drawing process. Fusing the drawing and line segment fitting in a single step saves time and improves the overall quality of the detected segments.
In addition, our proposal allows to jump over gradient discontinuities and detect full lines or just detect the individual linear segments without jumping.
This is important because line segments are features that, at the gradient level, can be easily broken by occlusions, shadows, glitches, etc.
In this way, the user can define the type of segments that best suits the application. For example, we may choose to detect large segments if the goal is to do Vanishing Points estimation or short ones for reconstruction and matching. 

\begin{figure}
    \centering
    \includegraphics[width=0.5\textwidth]{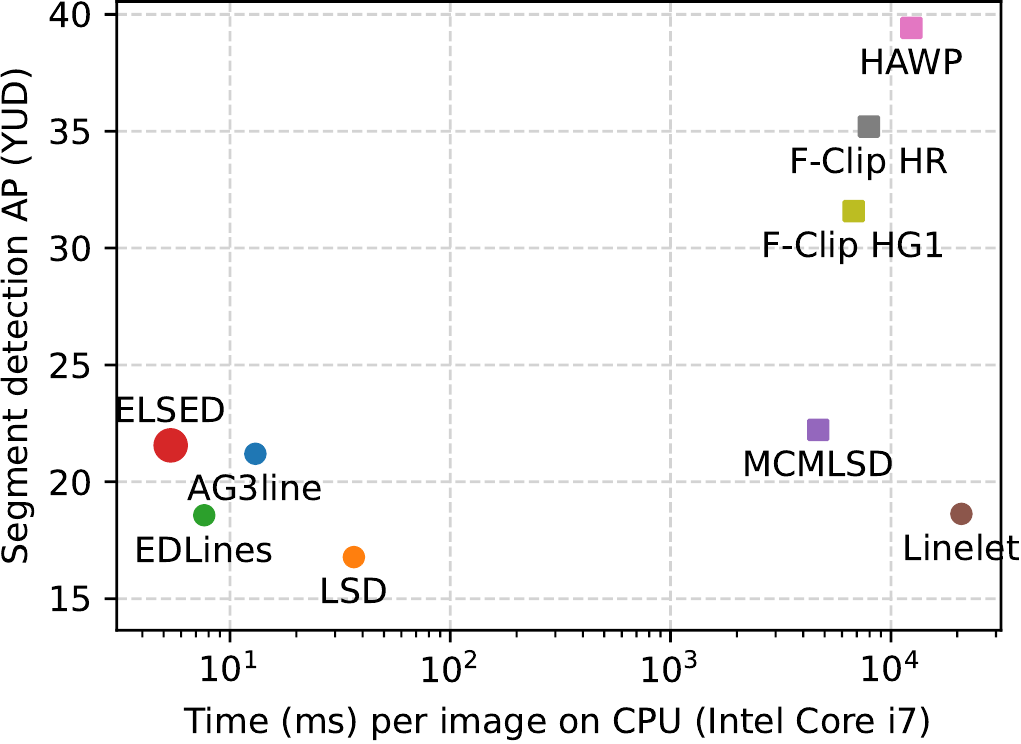}
    \caption{Average Precision (AP) vs. execution time (ms) curve in the line segment detection problem. Local features based methods are displayed with circular markers and global ones with square markers.} 
    \label{fig:sample_of_line_segments}
\end{figure}

In this paper we present an efficient method for line segment detection termed Enhanced Line SEgment Drawing (ELSED). In our experiments we compare the accuracy and efficiency of ELSED with that of the most relevant detectors in the literature. As shown in Fig.~\ref{fig:sample_of_line_segments}, ELSED is not only the most efficient (note the logarithmic scale in the speed dimension) but also the most accurate in line segment detection and more repeatable among the fastest in the literature, as we show in the experimental section.
%In our experiments we prove that ELSED provides best accuracy of the efficient segment detectors in the literature (see Fig.~\ref{fig:sample_of_line_segments}). %
%
%ELSED provides a new operation point in the accuracy vs resources trade-off curve that 
It improves the efficiency of present methods in resource-limited devices, opening the door to new CV applications running on any type of hardware. 

\section{Previous work}
\label{sec:state_of_art}

In this section we review the segment detection literature. To this end we organize it in three broad groups: full line detectors with global methods~\cite{matas2000robust, tal2012accurate, almazan2017mcmlsd}, those that use local properties to greedily detect line segments~\cite{grompone2010lsd, akinlar2011edlines, Ding2016OTLines, cho2017novel, zhang2021ag3line} and the deep line segment detectors~\cite{huang2018learning, xue2019learning, zhou2019end, xue2020hawp, dai2021fully, xu2021line, pautrat2021sold2}.

\subsection{Global information-based approaches} \label{sec:full_line_soa}

Global methods are able to detect full lines in the image with enough edge pixels support in spite of discontinuities. These methods start with an edge detection, for example Canny~\cite{canny1987computational}, and then they apply a Hough Transform-like~\cite{ballard1987generalizing} voting scheme. % duda1972use 
There are well-known issues with these methods: the omission of some weak edges, the generation of false positives in regions with high edge density (e.g. tree leaves) or the large amount of memory required to store the accumulators.

Progressive Probabilistic Hough Transforms (PPHT)~\cite{KIRYATI1991303, matas2000robust} solves the efficiency problem changing the entire image voting scheme by a random sampling scheme. 
%after this vote, the algorithm checks if the count be due to random noise, if it is not, a new line is detected and its pixels are removed from the voting ones. Moreover the line acceptance threshold is dynamically controlled. 
More recent works~\cite{tal2012accurate} address the quantization problem and also allow an efficient execution of the method. 
In \cite{XU20154012} the direction, length, and width of a line segment is extracted in a closed-form that uses a fitted quadratic polynomial curve. %
To generate better segments, MCMLSD~\cite{almazan2017mcmlsd} % is a line segment detector based on the Probabilistic Hough Transform~\cite{tal2012accurate}. The edge detector they
uses the Elder \& Zucker edge detector~\cite{elder1998edge} 
that propagates edge uncertainty to the Hough histogram accumulator. 
%In the final step 
Last, they split the detected full lines in line segments using a Markov Chain Model and a standard dynamic programming algorithm. % They also validate the resulting segments with a distribution learnt in the York Urban Database (YUD)~\cite{denis2008efficient}. % The interesting point of this method is that the Markov model is learnt from the annotated segments in the York Urban Database (YUD)~\cite{denis2008efficient}. 
% An important merit of this work is being one of the first that proposed an quantitative evaluation of the performance w.r.t. manually annotated segments in a number of images. 
%We use, among others, their evaluation measures in our experiments. 
%
%
The main drawback of this method is the efficiency. It takes 3.7 seconds to process a 640$\times$480 image. 
%
%An important merit of this work is being one of the first that proposed an quantitative evaluation of the performance w.r.t. manually annotated segments in a number of images. 
%This opens the door to a new Line Segment Detector that using the information available in the modern data sets and a empirical evaluation process, reach both: a high accuracy and a low computational cost.

%\subsection{Efficient local Segment Detectors: LSD and EDLines}
\subsection{Local methods} \label{sec:local_properties_soa}

Local methods overcome some of the drawbacks of global approaches by departing from strong gradient pixels and greedily add neighboring pixels using the gradient information. %These algorithms are able to return one pixel width continuous segments without the problems of full line approaches.
%
%
% LSD
%
LSD~\cite{grompone2010lsd} groups and validates image regions with a significant gradient magnitude and a similar gradient orientation in $O(N)$, being $N$ the number of image pixels. Unlike the non-maximal suppression (NMS) used in the Canny edge detector, LSD uses a region growing process to select interesting pixels. Then, 
each region is validated based on the expected Number of False Alarms (NFA), computed with an A-Contrario statistical model%
%proposed by Desolneux \emph{et al.}
~\cite{desolneux2000meaningful}. 
%
%This validation method finds segments as outlier alignments of pixels to a background statistical model of gradient orientations built over a Gaussian white noise image. 
LSD is  efficient and it is able to deal with areas with a high density of edge pixels (e.g. trees). %However, as pointed out by~\cite{akinlar2011edlines}, the validation step mainly rejects short segments, since long segments are always outliers to the background statistical model. We reach the same conclusion in the experiments in Section~\ref{sec:experiment-validation}.

%
% EDLines
%
EDLines~\cite{akinlar2011edlines} is also an efficient  algorithm ($O(N)$) based on local gradient orientations. It performs segment detection in two steps: 1) edge detection and 2) line segment detection using a local approach. The edge detection step is performed with the Edge Drawing (ED) algorithm~\cite{topal2012edge}. ED 
%is an alternative to the Canny edge detector that 
applies the first three Canny steps: Gaussian filtering, gradient estimation, NMS and tries to connect the anchor pixels (local maxima in gradient magnitude) with a greedy procedure. 
%ED is faster than Canny and returns a list of connected edge pixel chains instead of an edge map.
In a second step EDLines performs line fitting. % on each of these chains of pixels. Finally, it applies the same validation as LSD, although the authors state that this last step is optional. 
OTLines~\cite{Ding2016OTLines} uses an orientation transformation to improve EDLines and avoid segment detections on circular structures. % but has a slower implementation. 
AG3line~\cite{zhang2021ag3line} instead of drawing over all pixels only finds aligned anchors. 
%Thus, it performs in one step the two of EDLines (drawing and line fitting).
They also implement a continuous validation strategy to decide whether the segment has reached its endpoint and a jumping scheme to overcome gradient discontinuities. A key difference between AG3line and ELSED is that we use every pixel in the image as part of the drawing process, which generates a chain of continuous pixels, while AG3line uses only the detected anchors. This makes the method fast but unstable, needing to validate each step and thus being slower than ELSED.

% Linelet
%
The approach of Cho \emph{et al.}~\cite{cho2017novel} is based on \emph{linelets} detection, i.e. chunks of horizontally or vertically connected pixels that result from line digitization. The linelet detection is $O(N^2)$ time complexity and thus it takes 16.7 seconds per image. %, is conducted by first computing the gradient magnitude and orientation on every pixel, then performing NMS in the vertical and horizontal directions
%(removing non-maxima gradient gradient magnitude pixels) 
% and finally grouping the pixels in horizontal or vertical chains of connected pixels. 
Adjacent linelets are further grouped into line segments using a probabilistic model with $O(L^2)$ complexity, where $L$ is the number of detected linelets. Last, they validate using a mixture of experts model learnt from the gradient magnitudes, orientation and from the line length of the segments in a labeled data set. 
They also propose a quantitative evaluation that we improve in our experiments (section~\ref{sec:experiments}).

\subsection{Deep line segment detectors}\label{sec:soa_endpoint_prediction}

A closely related problem to line segments detection is wireframe parsing. It consists of predicting the scene's salient straight lines and their junctions. The ShanghaiTech Wireframe data set~\cite{huang2018learning} contains over 5,000 hand-labelled images that allowed different methods to train obtaining competitive results. 
AFM~\cite{xue2019learning} uses an attraction field map that is next squeezed to obtain line segments. 
L-CNN~\cite{zhou2019end} proposes an End-to-end model that uses a stacked hourglass %~\cite{newell2016stacked} 
backbone to obtain a junction proposal heatmap that is extensively sampled to obtain the output segments. 
HAWP~\cite{xue2020hawp} improves the L-CNN sampling step by reparametrizing the line segments in a holistic 4-D attraction field map from which segments can be obtained faster. 
HT-HAWP~\cite{xue2020hawp} Adds some Deep Hough layers to the HAWP model improving its capabilities to capture lines and slightly improving the performance in some benchmarks.
F-Clip~\cite{dai2021fully} proposes a simple yet effective approach to cast line segment detection as an object detection problem that can be solved with a fully convolutional one-stage method. %It also shows how modifying the backbone and exploiting multiple resolutions of the input through a parallel structure can vary the trade off between speed and accuracy. 
LERT~\cite{xu2021line} detects segments using transformers that replace the junction heatmap and segment proposals to directly predict the segment endpoints.
$\text{SOLD}^2$~\cite{pautrat2021sold2} proposes a self-supervised way towards line detection and description that optimizes the repeatability of the detected segments.

Despite the good results of these deep methods, their computational requirements are still far away from the classical methods based on gradient. This fact makes them non-viable for limited devices where there is no GPU or rather its battery consumption is prohibitive.
For this reason we introduce our drawing method that has been carefully designed to avoid floating-point operation and minimize its complexity, being CPU friendly and achieving the fastest execution times on the state of the art.

\section{Enhanced Line SEgment Drawing (ELSED)}
\label{sec:smart_edlines_implementation}

In this section we introduce our line segment detection method %. The EDLines algorithm for segment detection~\cite{akinlar2011edlines} splits the detection in two steps: 1) edge pixels detection and 2) line segment detection. First, we have realized that these two steps in the EDLines can be performed at the same time. Second, we introduce Enhanced Edge Drawing (EED), an algorithm that is more efficient than the original Edge Drawing algorithm and also improves the edges obtained in some cases. 
%Third, we have enhanced the ED routing algorithm taking into account that we are looking specifically for line segments.  In this section we 
and explain the different steps in our approach. 

\subsection{Enhanced Edge Drawing algorithm (EED)}
\label{sec:enhanded-edge-drawing-algorithm}

The EED entails the following high-level steps:
1) Gaussian smoothing to suppress noise;
2) Gradient magnitude and orientation computation;
3) Extraction of anchor pixels, local maxima in the gradient magnitude;
4) Connect the anchors using the enhanced routing algorithm

For the noise reduction step we use a convolution with a $5\times 5$ Gaussian kernel and $\sigma=1$, for Gradient magnitude and orientation computation we first compute the horizontal, $G_x$, and vertical, $G_y$, gradients by applying the Sobel operator and then we use the $L1$ norm, $G=|G_x| + |G_y|$. We also define a gradient threshold and set $G=0$ for those pixels below it and quantize the gradient orientation, $O$, into two possible values: vertical edge, $|G_x| \ge |G_y|$, or horizontal edge, $|G_x| < |G_y|$. The other two steps are explained in the next subsections. 

\subsubsection{Extraction of anchor pixels}
\label{sec:anchors}

The anchors are pixels where the drawing process begins. %They may not be included in the set of final line segments if they belong to a very small. 
%
%We detect anchors in the same way as in the ED original algorithm. 
%Like in ED, 
We scan image pixels with $G>0$ and test if it is a local maxima in the gradient magnitude, $G$, along the quantized direction of the gradient, $O$.
If the pixel orientation $O(x, y)$ corresponds to a vertical edge, it is an anchor if $G[x,y]-G[x-1, y] \ge T_{anchor}$ and $G[x,y]-G[x+1, 
y] \ge T_{anchor}$. The same applies for horizontal edge in vertical direction.
%On the other hand, if the pixel orientation corresponds to an horizontal edge, it is an anchor if $G[x,y]-G[x, y-1] \ge T_{anchor}$ and $G[x,y]-G[x, y+1] \ge T_{anchor}$. 
To increase the processing speed, the number of anchors can be limited by increasing the value of $T_{anchor}$ and 
also by scanning pixels every $SI=2$ columns and rows. 
%(i.e. not all pixels are tested for the local maxima condition).

\subsubsection{Connecting the anchors by an enhanced routing algorithm}
\label{sec:drawing-algorithm}

\begin{figure}
    \begin{subfigure}[b]{0.45\textwidth}
    \centering
    \footnotesize
    \includegraphics[width=\textwidth]{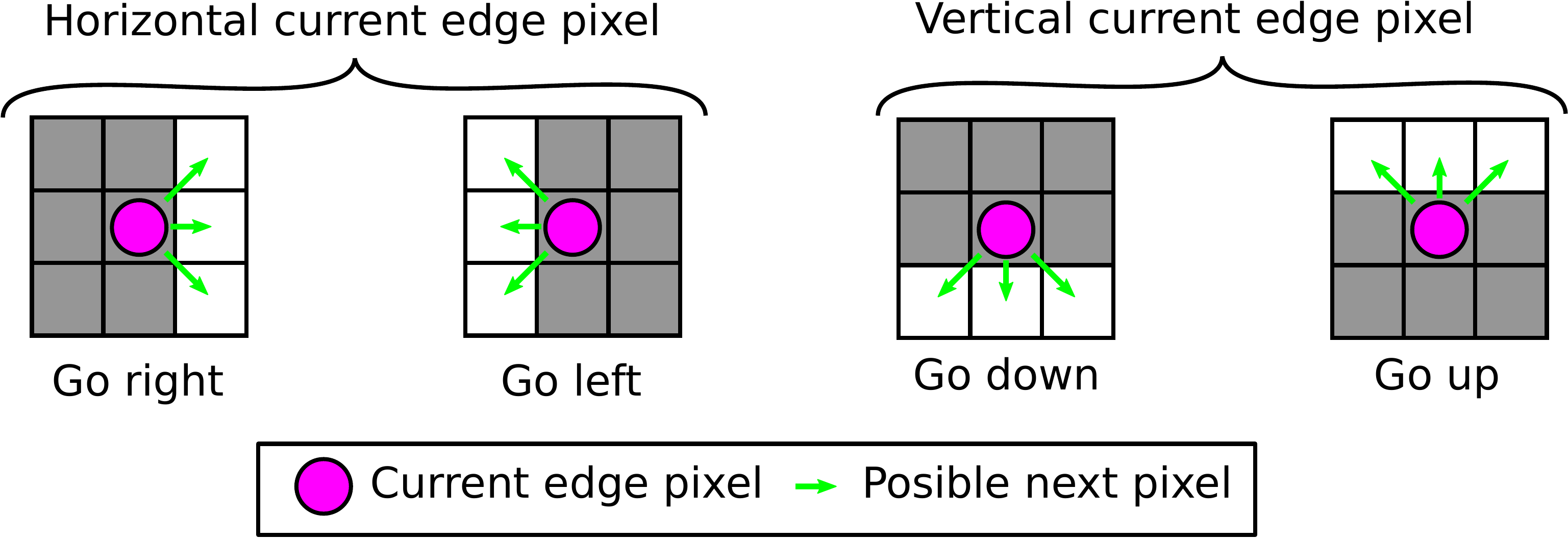}
    \caption{ED next pixel selection.}
    \label{fig:edge-drawing}
    \end{subfigure}
    \begin{subfigure}[b]{0.5\textwidth}
    \centering
    \footnotesize
    \includegraphics[width=\columnwidth]{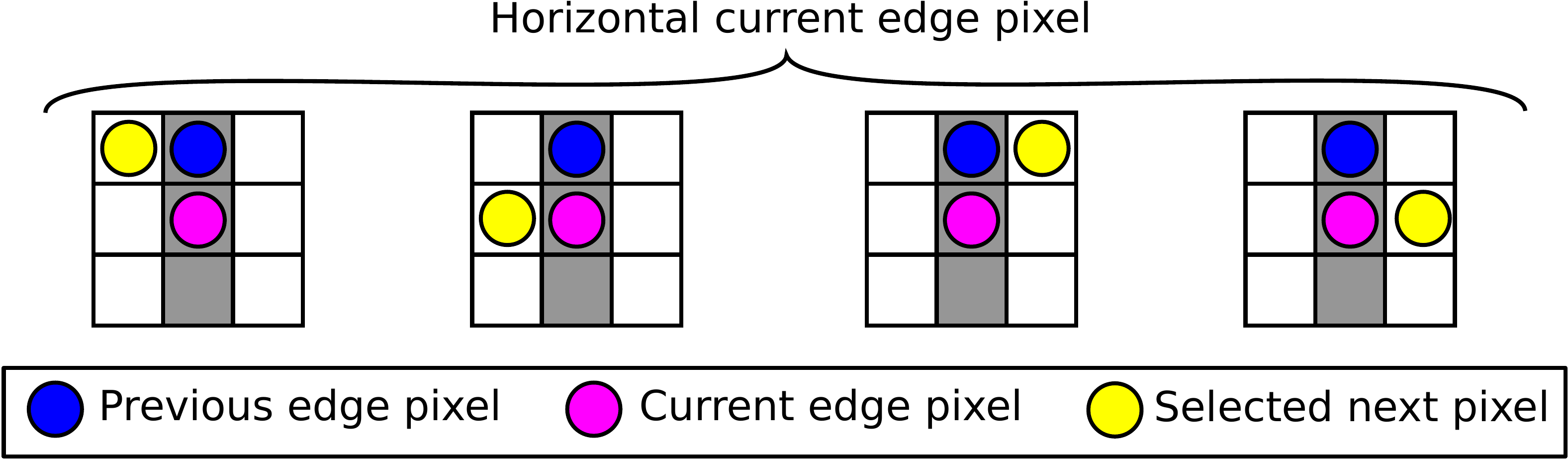}
    \caption{Corner-shaped edges arising from ED.}
    \label{fig:edge-drawing-problems-corners}
    \end{subfigure}
    \caption{ED greedy segment growing. In (a) it only takes into account the current edge pixel. In the case of (b) the walk is coming from the blue pixel and finds a horizontal edge pixel (pink one). Thus, it will start a walk to the left and another to the right that could give edge chains with the displayed configurations (blue-pink-yellow pixel sequence).}
\end{figure}

ED is faster than LSD's region growing because, from an initial anchor, it only walks along a chain of edge pixels, evaluating 3 neighbours at each step (see Fig.~\ref{fig:edge-drawing}) and selecting as next step the one with biggest gradient magnitude. The evaluations are critical for the speed of the algorithm since they are done for each reachable edge pixel walking from an anchor point. 

\begin{figure}
    \centering
    \footnotesize
    \begin{subfigure}{0.49\textwidth}
        \centering
    \footnotesize
        \includegraphics[width=\columnwidth]{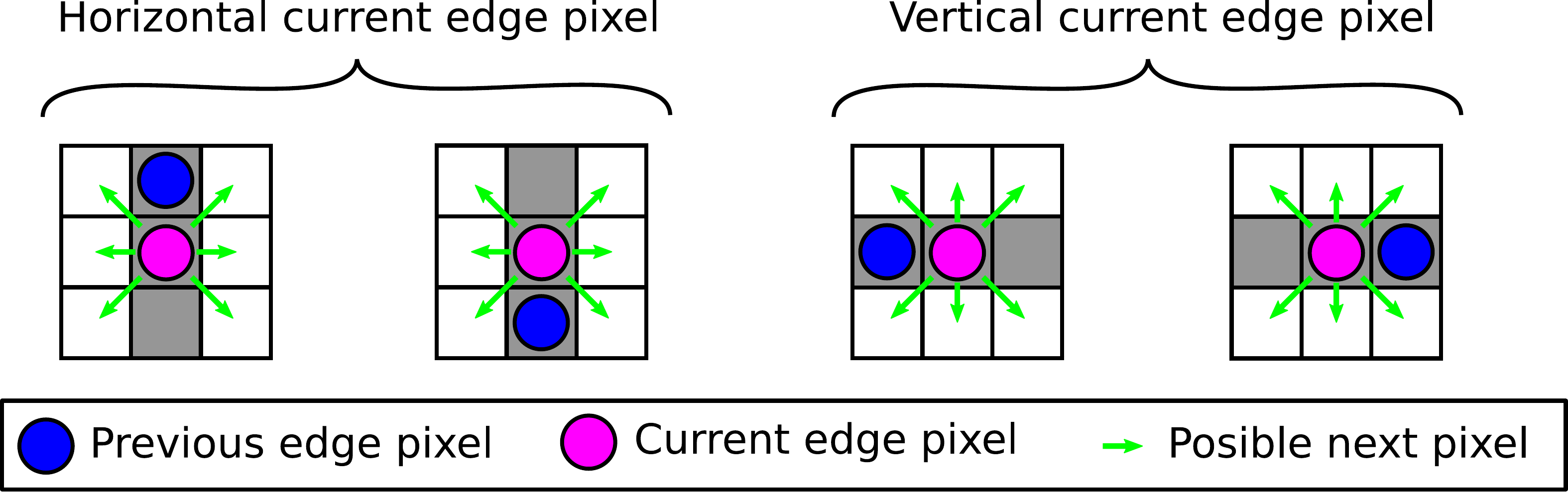}
        \caption{Edge Drawing (ED).
        }
        \label{fig:edge-drawing-problems-previous-pixel}
    \end{subfigure}
    \begin{subfigure}{0.49\textwidth}
        \centering
    \footnotesize
        \includegraphics[width=\columnwidth]{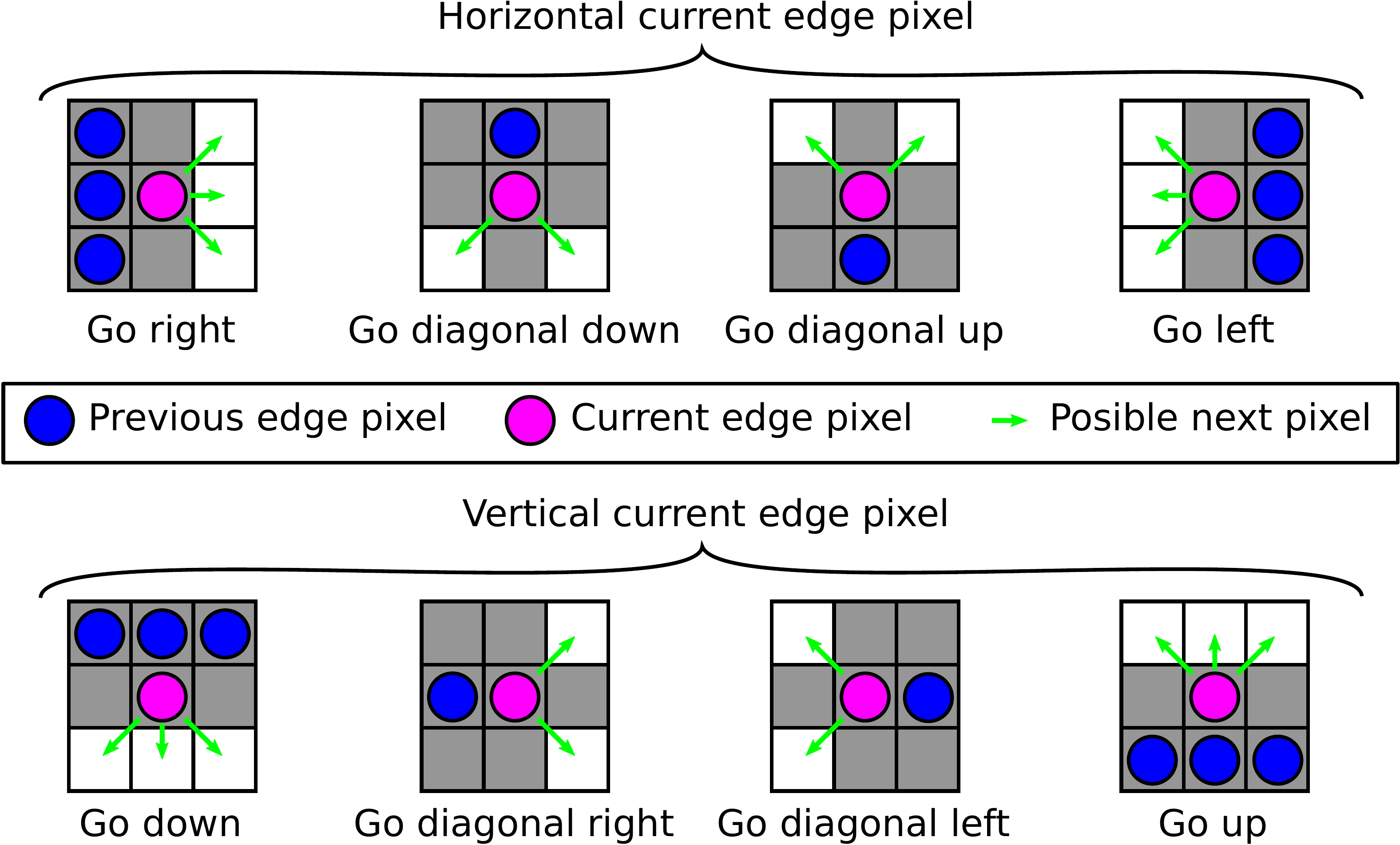}
        \caption{Enhanced Edge Drawing (EED). 
        }
        \label{fig:enhanced-edge-drawing}
    \end{subfigure}
    %\caption{Original ED vs EED candidate pixels when the previous pixel was in a different orientation (vertical vs horizontal) than the current one. EED reduces the number of explored pixels from 6 to 3 or 2 by taking into account the line segment direction.
    %}
    \caption{Drawing diagonal edges: When the previous pixel was in a different orientation (vertical vs horizontal) than the current one, original ED has 6 candidates pixels vs EED that has 3 or 2. %EED reduces the number of explored pixels from 6 to 3 or 2 by taking into account the line segment direction. 
    More than one previous edge pixel is displayed when we have the same candidates for any of the previous pixel options.
    }    
    \end{figure}

In our EED procedure, we perform the edge drawing and line fitting at the same time. This will enable us to save computations by reducing the number of checked pixels. 
% In each step of the walking process, once we know that we are following a segment, we restrict the candidates to the next pixel to those that follow the actual line. 
%
During drawing, we consider the previous and the current pixel. %, and not only the walking direction as in ED (i.e. right, left, up or down). 
We explore the same pixels as in ED during the walking processes (see ``Go right'', ``Go left'', ``Go up'' and  ``Go down'' in Fig.~\ref{fig:enhanced-edge-drawing}) %than in the ED algorithm 
as long as the edge orientation does not change from the previous pixel to the current one. When the previous pixel is in a vertical edge, and the current pixel is in a horizontal one, ED has 6 candidate pixels to be added to the current line segment (see Fig.~\ref{fig:edge-drawing-problems-previous-pixel}). %The other way around is also true, when previous pixel is in an horizontal edge and current one is on a vertical one, ED has 6 candidate pixels.
This may generate situations where the algorithm would draw a corner breaking the line assumption (Fig.~\ref{fig:edge-drawing-problems-corners}). 
The same happens when the previous pixel is in a horizontal edge and the current is on a vertical one.

Here we introduce a different approach to treat these diagonal pixels while following a line. We add the assumption that the edge chain should form a line. Then, considering Bresenham's line drawing scheme, the number of checked pixels in this situation changes from 6 with ED (see Fig.~\ref{fig:edge-drawing-problems-previous-pixel}) to only 2 with EED (see the four ``diagonal'' cases in Fig.~\ref{fig:enhanced-edge-drawing}), and remains 3 for the other possible previous pixels (non-diagonal cases). This has two advantages: 1) it is faster than the original ED routing algorithm as it explores fewer pixels and, 2) it avoids non-meaningful cases for finding line segments. 

\begin{figure}
    \centering
    \begin{subfigure}{0.3\textwidth}
     \centering
     \includegraphics[width=\textwidth]{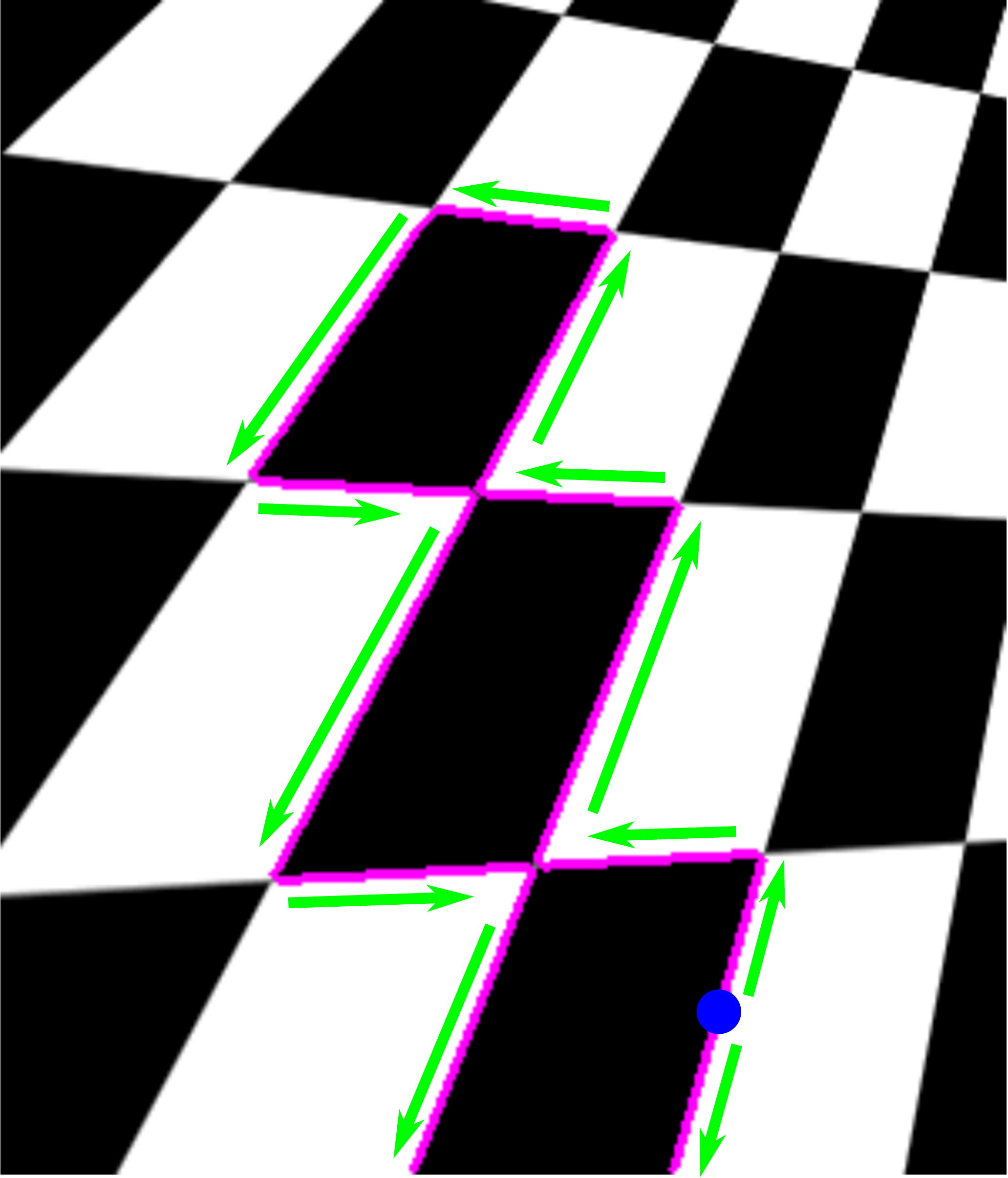}
     \caption{Edge Drawing~\cite{topal2012edge}}
     \label{fig:drawing-result-1}
    \end{subfigure}
    \begin{subfigure}{0.3\textwidth}
     \includegraphics[width=\textwidth]{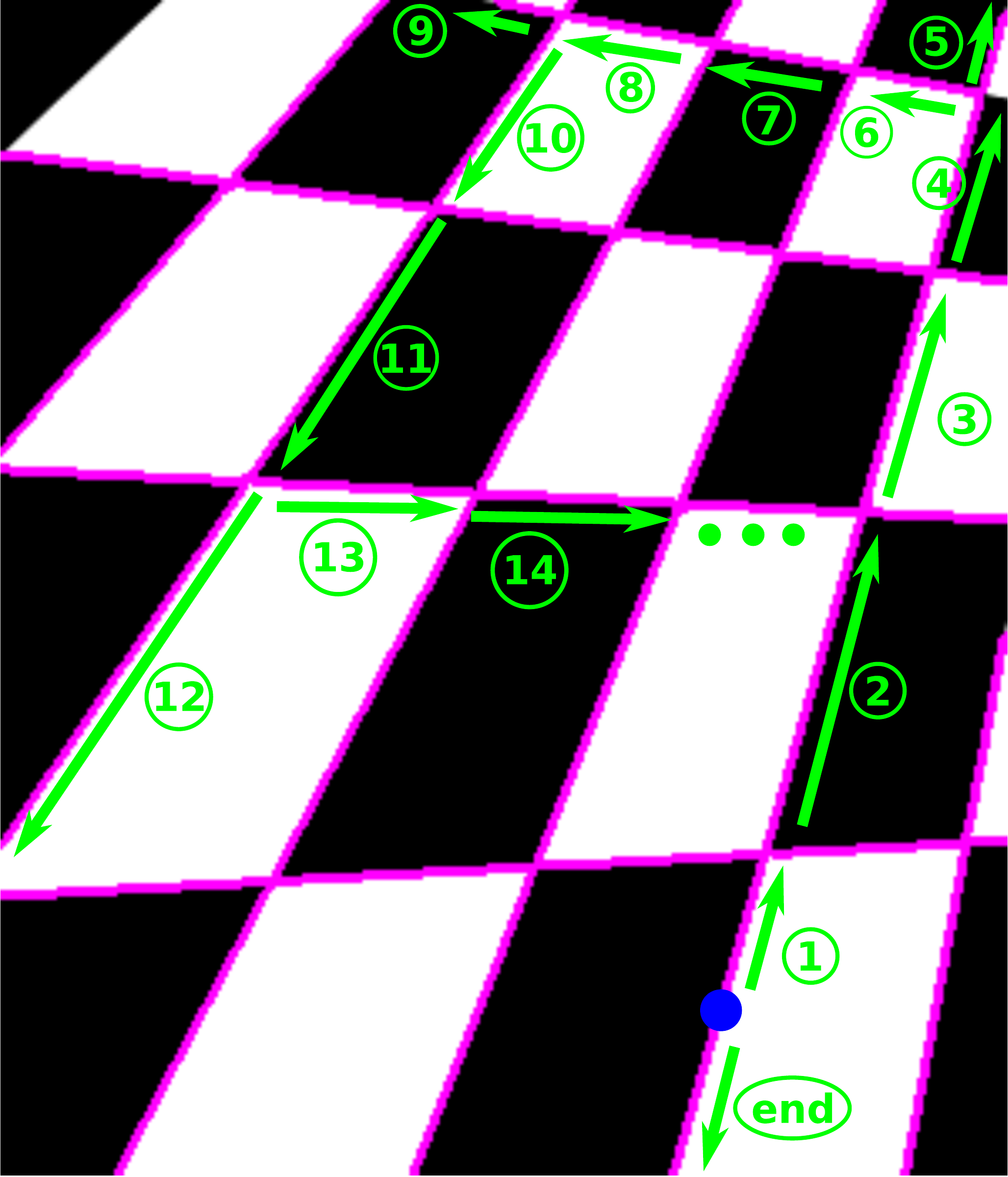}
     \caption{ELSED}
     \label{fig:drawing-result-2}
    \end{subfigure}
     \caption{Results of the  drawing process for ED and EED (purple pixels) from one single anchor (blue point). Green arrows and numbers establish which segments are visited first. %Since our method is aware that it is following a line it is able to continue drawing when the edge orientation changes. 
     %ED on the other side, follows the maximum gradient magnitude edge pixels breaking long line segments into short ones and missing some of them.
     }
     \label{fig:drawing-result}
\end{figure}

The second important idea is a also consequence of trying to find aligned edge pixels. Whereas ED changes the walking process direction when a change of edge orientation is detected (See Fig. \ref{fig:drawing-result-1}), EED tries to continue in the same direction following a line. However, any change of edge orientation is not forgotten and it is pushed into the stack of discontinuities, $\mathcal{D}_{stack}$, for later processing (see Algorithm~ \ref{alg:drawing-stack-algorithm}).
%During the drawing process, the perpendicular gradient direction is followed using EED. 
%
EED tries to fit a line to the current chain of pixels, if more than $T_{minLength}$ pixels have been chained and the squared error of alignment of the pixels is lower than $T_{LineFitErr}$. The last parameter for the segment search is the $T_{PxToSegDist}$ that is the maximum distance in pixels from which we consider whether a pixel fits or not in the current segment. This is done internally in the function \emph{addPxToSegment} in line \ref{alg:addpxToSegment} of Algorithm~\ref{alg:drawing-stack-algorithm}).
The process stops when no more pixels can be chained (i.e. at the limits of the image, with only already visited pixels or weak edge pixels as candidates) or a line edge discontinuity is detected (e.g. the gap between two aligned windows, a tree branch occluding part of a building, etc.). In case a discontinuity was detected we execute, in order, the following actions:
\begin{enumerate}
    \item[1.] If we were walking along a line segment (i.e. we have fitted a line), try to extend it going on the line direction (the walking process stacked using \emph{canContinueForward} and \emph{forwardPxAndDir} functions in algorithm~\ref{alg:drawing-stack-algorithm}, lines \ref{alg:canContinueFordward} to \ref{alg:endCanContinueFordward}) 
    \item[2.] If were walking along a line segment and we cannot continue forward, try to extend the line segment backwards (the walking process stacked using \emph{canContinueBackward} and \emph{backwardPxAndDir} functions in algorithm~\ref{alg:drawing-stack-algorithm}), lines \ref{alg:canContinueBackward} to \ref{alg:endCanContinueBackward})
    \item[3.] Continue in the gradient direction, that is changing in the discontinuity (the walking process stacked in Alg.~\ref{alg:drawing-stack-algorithm}, line~\ref{alg:drawOtherDirection}).
\end{enumerate}

This sequence of ordered actions guarantees that if a line segment is detected, all its pixels will be detected together. 
% hence, our segment structures are little more than pointers to the list of pixel edges.

We show an illustrative example in Fig.~\ref{fig:drawing-result-2}, where EED starts two walking processes, one upward and another downward. Unlike ED (see Fig.~\ref{fig:drawing-result-1}), EED detects the discontinuities in the edge orientation of the chessboard corners and continues walking in the current line direction. 
Each discontinuity is stored in $\mathcal{D}_{stack}$ for later processing.
After pixels in segments (1) to (5) are chained, 
%the last edge orientation change is extracted from the top of $\mathcal{D}_{stack}$. 
the next edge orientation to process is extracted from the top of $\mathcal{D}_{stack}$.
The drawing process keeps drawing from it, linking the pixels of segments (6) to (9). The routing algorithm ends when there are no more edge orientation changes in $\mathcal{D}_{stack}$. Then, the next anchor point is processed by the routing algorithm. 
Detecting more segments from a single anchor is an important feature of our method that increases the detection recall with respect to EDLines as we will show in experiment~\ref{sec:exp-seg-detection}.

\begin{algorithm}
	\caption{Enhanced Edge Drawing Algorithm}
    \label{alg:drawing-stack-algorithm}
    \scriptsize
    \linespread{1}
\begin{algorithmic}[1]
\Procedure{EED}{$a, d_0$}\\ 
	\textbf{Input:} Anchor pixel: $a$. Anchor gradient direction: $d_0$\\ 
	\textbf{Output:} S: List of Segments, P: List of edge pixels
	\State $P \leftarrow \{a\}$ ; $S \leftarrow \emptyset$
	\State $\mathcal{D}_{stack}$ $\leftarrow \emptyset$ ; $\mathcal{D}_{stack}$.push([$a$, $d_0$])
	\While{$\mathcal{D}_{stack}$ $\neq \emptyset$} %$M \neq \emptyset$}
		\State segmentFound $\leftarrow$ \textbf{false}
		\State nOutliers $\leftarrow$ 0
		%\State px $\leftarrow$ top($\mathcal{D}_{stack}$).anchor
		%\State lastPx $\leftarrow$ top($\mathcal{D}_{stack}$).anchor - top($\mathcal{D}_{stack}$).dir
		\State $c, d \leftarrow$ $\mathcal{D}_{stack}$.pop() \Comment Current pixel and direction
		\State $p \leftarrow$ previousPixel($c$, $d$) \Comment Find previous pixel
		\While{$G[c] \neq 0$ $\wedge$ nOutliers $\leq T_{outliers}$ } %$M \neq \emptyset$}
			\State $c, p$ $\leftarrow$ drawNextPx($c$, $p$)
			\State $P \leftarrow P \cup \{c\}$
			\If{segmentFound} 
				\State{ s, nOuliers $\leftarrow$ addPxToSegment(s, $c$, nOuliers)} \label{alg:addpxToSegment}
			\Else 
				\State s, segmentFound $\leftarrow$ fitNewSegment($P$) \Comment Can return $\emptyset$
				\State S $\leftarrow$ S $\cup$ \{s\}
			\EndIf
		\EndWhile
		\If{$G[c] \neq 0$} 
			\State $\mathcal{D}_{stack}$.push([$c$, lastDir]) \label{alg:drawOtherDirection} \Comment Edge orientation change
			\If{canContinueForward(s)} \label{alg:canContinueFordward}
                \State $p_f, d_f \leftarrow$ s.forwardPxAndDir($c$, $d$)
				\State $\mathcal{D}_{stack}$.push([$p_f$, $d_f$]) 
			\EndIf  \label{alg:endCanContinueFordward}
			\If{canContinueBackward(s)} \label{alg:canContinueBackward}
				\State $p_b, d_b \leftarrow$ s.backwardPxAndDir($c$, $d$)
				\State $\mathcal{D}_{stack}$.push([$p_b$, $d_b$]) 
			\EndIf \label{alg:endCanContinueBackward}
		\EndIf
	\EndWhile
\EndProcedure
\end{algorithmic}
\end{algorithm}

\subsection{The line segment discontinuities}
\label{sec:discontinuity_management}

A line segment can be interrupted by several edge discontinuities. These are regions of the image where the gradient orientation changes or the gradient magnitude goes to zero. Whether we aim to detect full lines or just line segments, the discontinuities of different lengths (i.e. the number of pixels where there is no edge or the edge direction is not aligned with the line segment) should be correctly skipped in a drawing process in order to detect the line segments correctly. 

Algorithm \ref{alg:drawing-stack-algorithm} naturally deals with this phenomena. %detecting discontinuities where the last pixels drawn do not fit in the current segment line. 
%This is done by using a threshold $PxToSegmentDistTh$. 
%
Once a discontinuity is detected, our aim is to jump over it and continue drawing in the line segment direction if possible. \emph{A priori}, the discontinuity length  is unknown and our algorithm test different length candidates. As we are using a $5\times5$ Gaussian smoothing kernel, any 1 pixel discontinuity will have effect in at least a neighbourhood of size 5 pixels, therefore we set the minimum discontinuity length to 5 pixels. In the functions \textit{canContinueFordward} (Alg. \ref{alg:drawing-stack-algorithm}, line~\ref{alg:canContinueFordward}) and \textit{canContinueBackward} (Alg. \ref{alg:drawing-stack-algorithm}, line~\ref{alg:canContinueBackward}), different jump lengths $J$ are checked (in the default parameters of the algorithm we use $J\in [5, 7, 9]$). The drawing process will continue after the discontinuity if the ordered conjunction of the following conditions is true:
\begin{enumerate}
    \item The segment is longer than the number of pixels, $J$, we want to jump.
    \item The pixel, $a$, that is aligned with the segment and $J$ pixels away from current pixel, $c$, is inside the image and has $G[a]>0$ (i.e. is not a weak edge pixel).
    \item Starting from $a$, EED is able to draw at least $J$ pixels following the edge direction. We call this set of $J$ pixels the extension pixels.
    \item The extension pixels are well aligned with the line segment. To check this we calculate the auto-correlation matrix of the image gradients, $\mM$, in a small neighbourhood (we take one pixel on each side of the extension pixels) and then, we assert that:
    \begin{equation}
        \frac{\lambda_1}{\lambda_2} \geq T_{EigenExt}
        \label{eq:discontinuities-eigen-cond}
    \end{equation}
    where $\lambda_1$ and $\lambda_2$, $(\lambda_1>\lambda_2)$, are the eigenvalues of the $\mM$, and 
    \begin{equation}
        \angle(\vv_1 , \vn ) \leq T_{AngleExt}
        \label{eq:discontinuities-angle-cond}
    \end{equation}
    $\angle(\cdot, \cdot)$ is the angular distance, $\vv_1$ is the first eigenvector of $\mM$ and $\vn$ is a vector normal to the segment.
\end{enumerate}

\begin{figure*}
    \centering
    %\footnotesize
    \begin{subfigure}{.33\textwidth}
     \centering
    \footnotesize
     \includegraphics[height=\columnwidth]{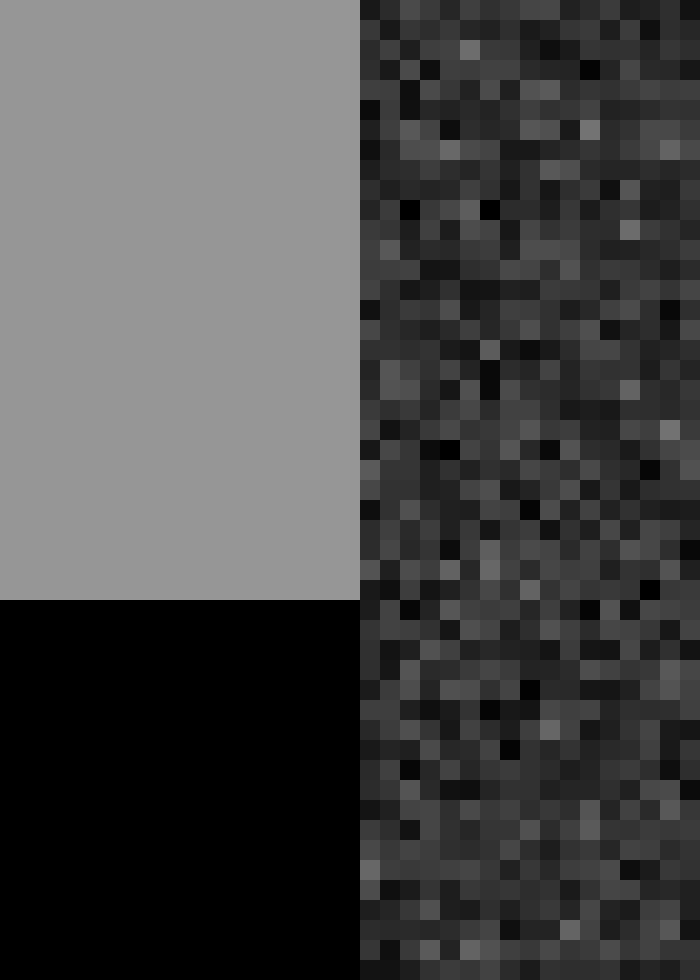}
     \caption{Original Image}
     \label{fig:discontinuity-management-original}
    \end{subfigure}%
    \vspace{0.5ex}%
    \begin{subfigure}{.33\textwidth}
     \centering
    \footnotesize
     \includegraphics[height=\columnwidth]{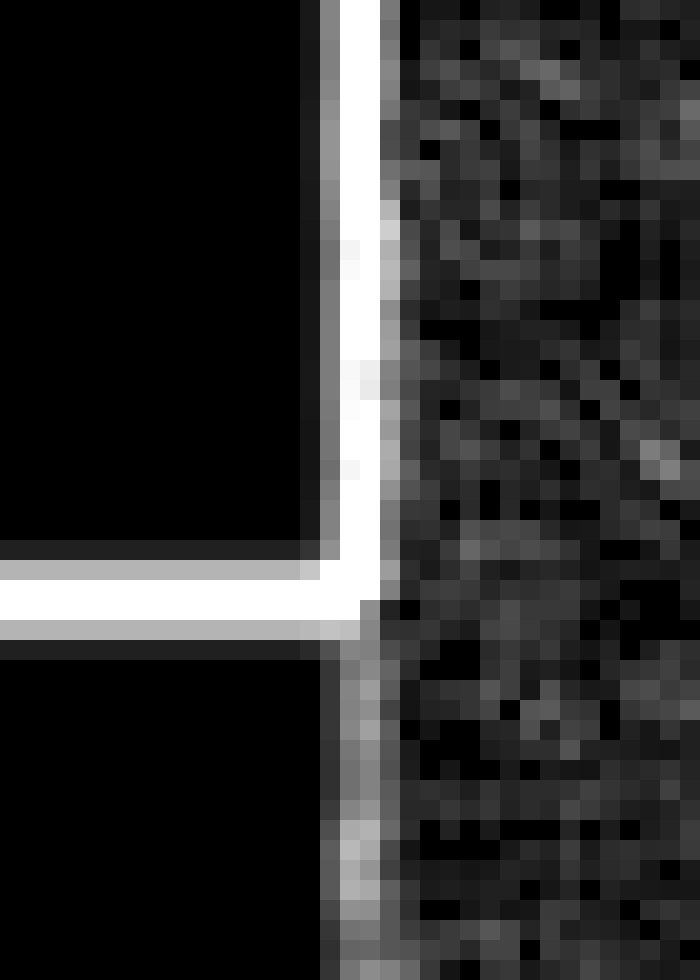}
     \caption{Blurred Image gradient}
     \label{fig:discontinuity-management-gradient}
    \end{subfigure}%
    \begin{subfigure}{.33\textwidth}
     \centering
    \footnotesize
     \includegraphics[height=\columnwidth]{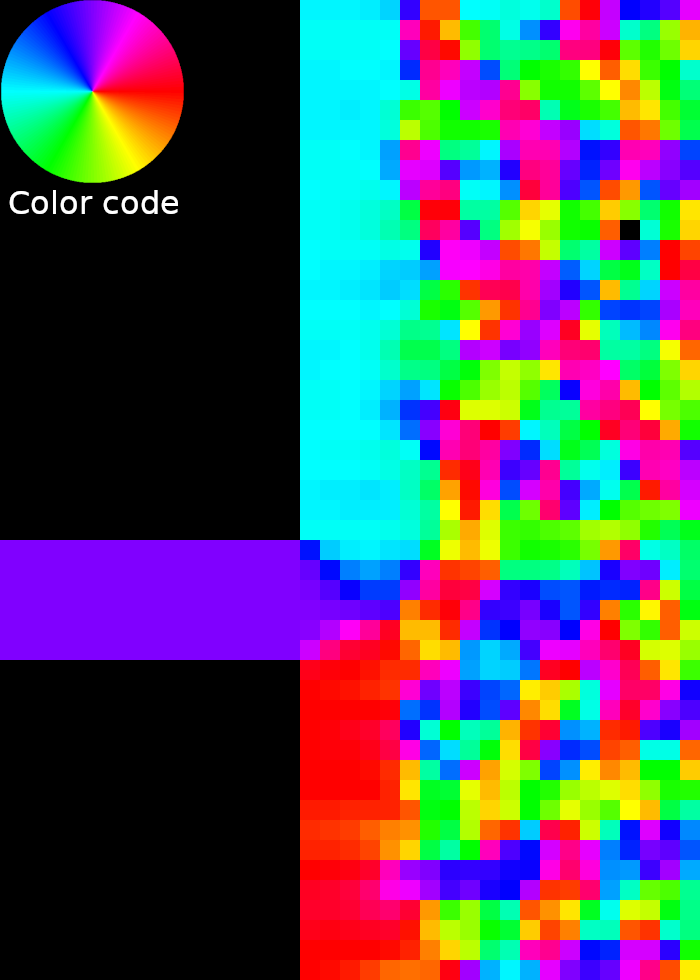}
     \caption{Gradient orientation}     
     \label{fig:discontinuity-management-gradient-orientation}
    \end{subfigure}
    \begin{subfigure}{.33\textwidth}
     \centering
    \footnotesize
     \includegraphics[height=\columnwidth]{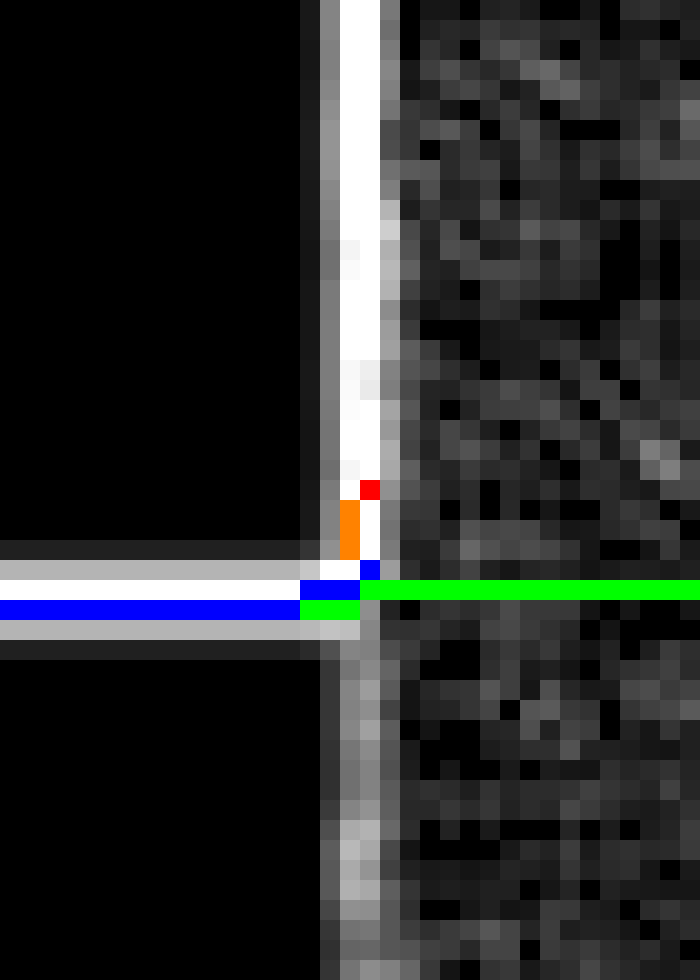}
     \caption{Discontinuity detection}
     \label{fig:discontinuity-management-outliers}
    \end{subfigure}%
    \begin{subfigure}{.33\textwidth}
     \centering
    \footnotesize
     \includegraphics[height=\columnwidth]{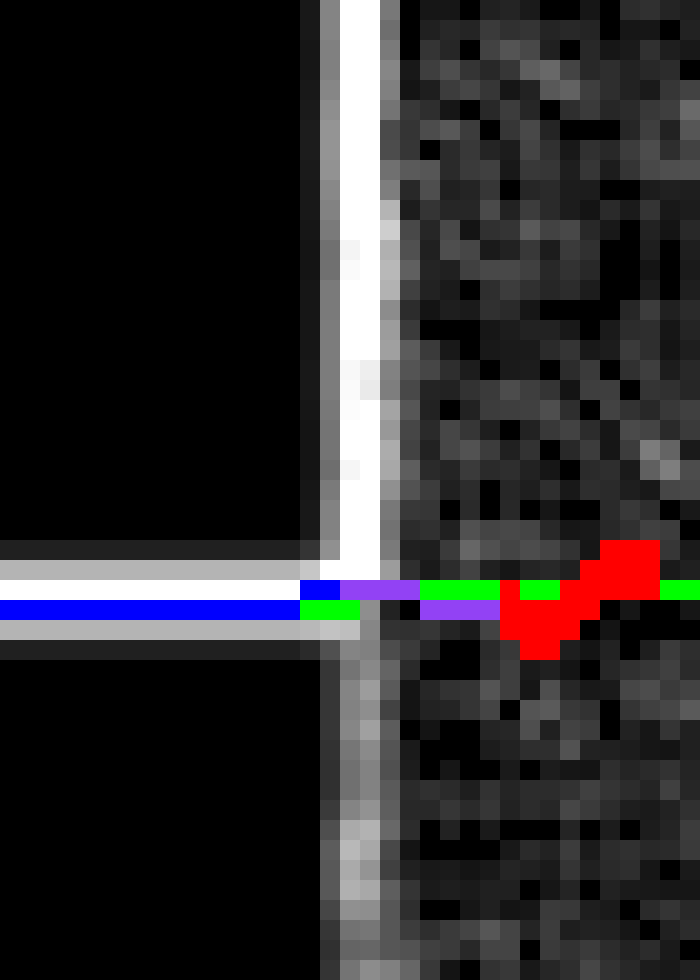}
     \caption{1st discontinuity check}
     \label{fig:discontinuity-management-extension1}
    \end{subfigure}%
    \begin{subfigure}{.33\textwidth}
     \centering
    \footnotesize
     \includegraphics[height=\columnwidth]{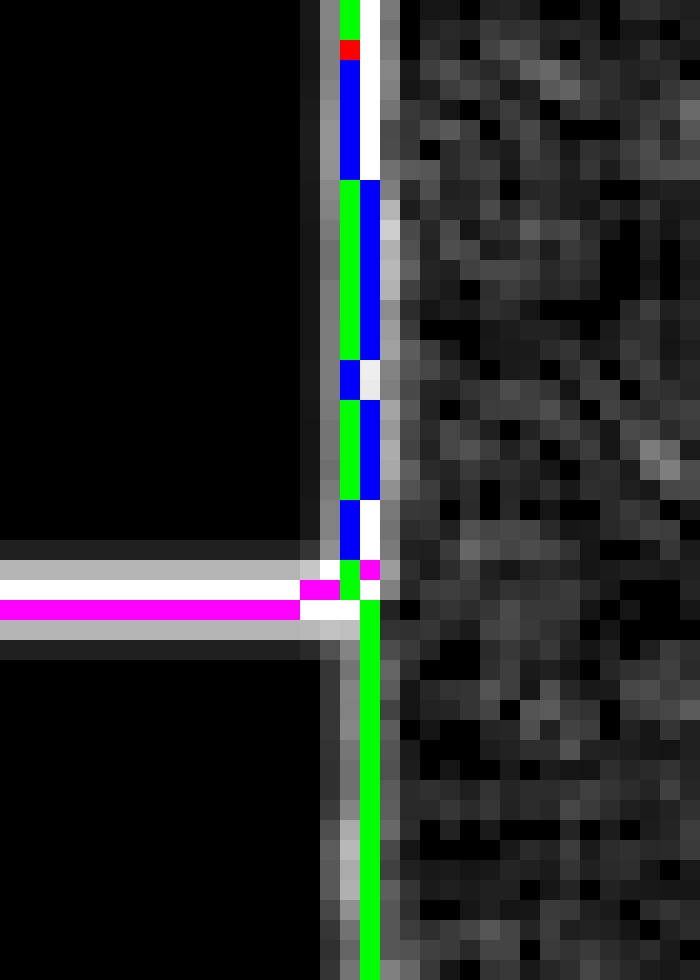}
     \caption{Drawing 2nd segment}
     \label{fig:discontinuity-management-going-up}
    \end{subfigure}
    \begin{subfigure}{.33\textwidth}
     \centering
    \footnotesize
     \includegraphics[height=\columnwidth]{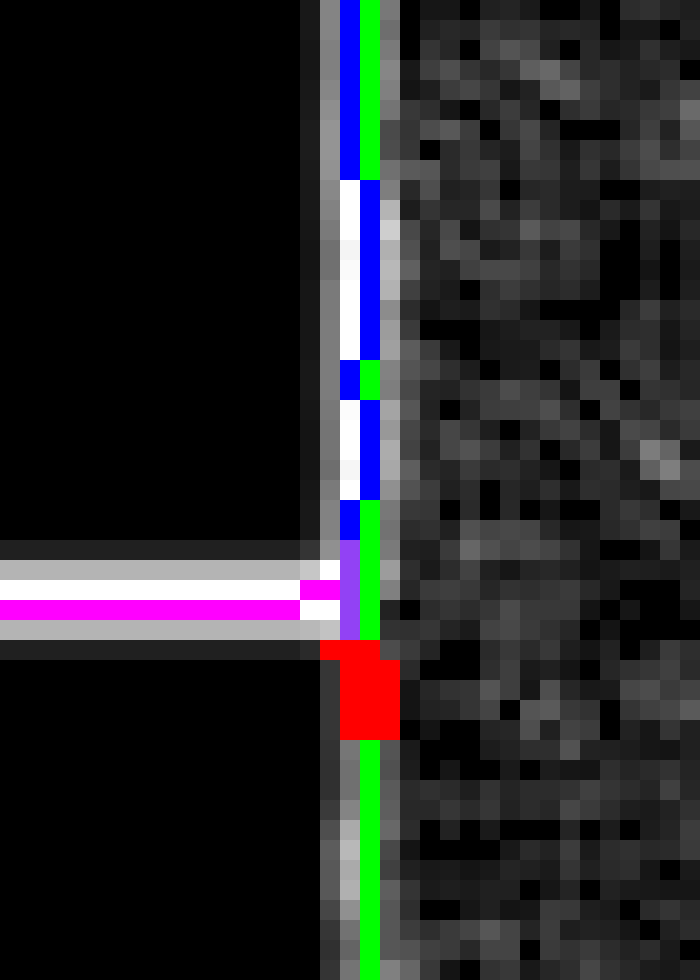}
     \caption{2nd discontinuity check}
     \label{fig:discontinuity-management-extension2}
    \end{subfigure}%
    \begin{subfigure}{.33\textwidth}
     \centering
    \footnotesize
     \includegraphics[height=\columnwidth]{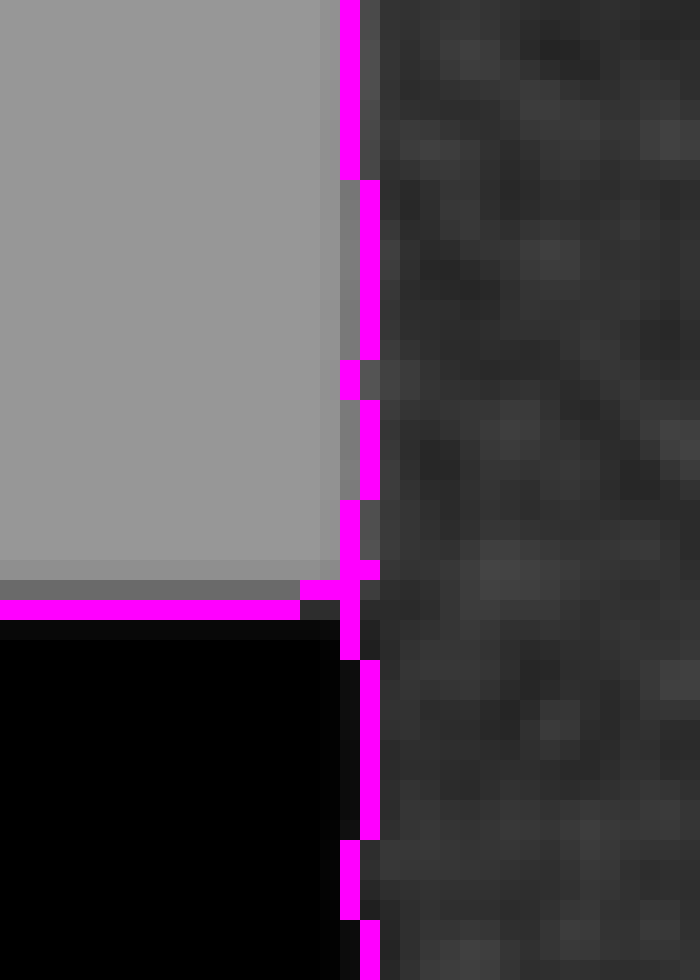}
     \caption{Detected edges}
     \label{fig:discontinuity-management-edges}
    \end{subfigure}%
    \begin{subfigure}{.33\textwidth}
     \centering
    \footnotesize
     \includegraphics[height=\columnwidth]{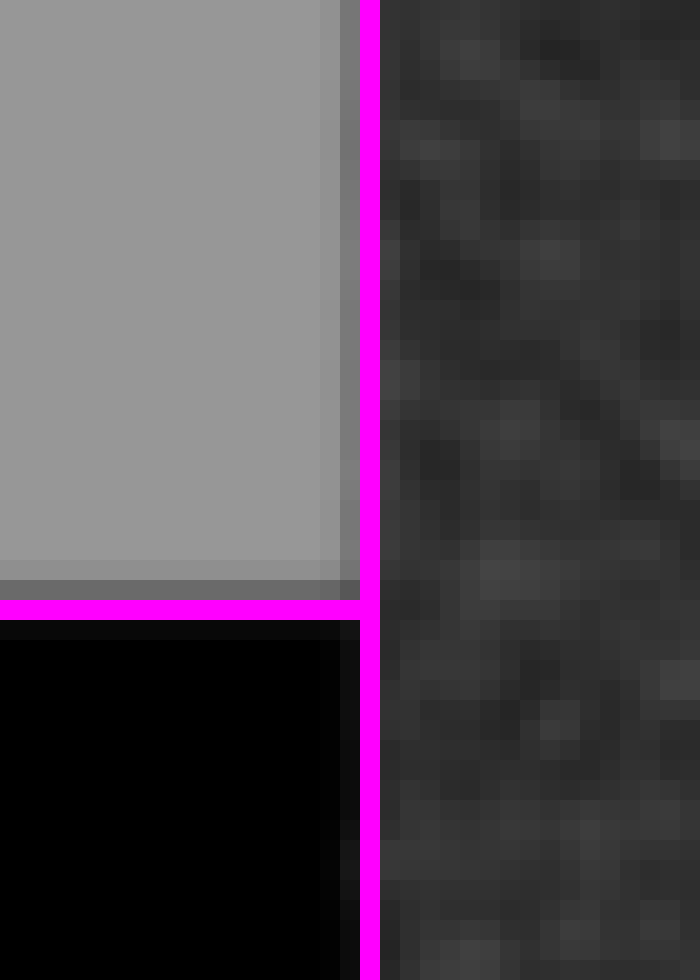}
     \caption{Detected Segments}
     \label{fig:discontinuity-management-segments}
    \end{subfigure}
    \caption{Example of ELSED discontinuity management algorithm. Gradient orientations in (c) coded with red (right direction), purple (up), cyan (left) and light green (bottom)}
    \label{fig:discontinuity-management}
\end{figure*}

In Fig.\ref{fig:discontinuity-management} we show different steps of the discontinuity management algorithm in a synthetic image. 
%Figs.~\ref{fig:discontinuity-management-original}, \ref{fig:discontinuity-management-gradient} and \ref{fig:discontinuity-management-gradient-orientation} show respectively, the original gray-scale image, the gradient of the blurred image and the gradient orientation where segments appear as uniform regions with the same gradient direction.
%Colors of the gradient orientation in Fig.~\ref{fig:discontinuity-management-gradient-orientation} go from red (right direction), purple (up), cyan (left) and light green (bottom). 
%
In Fig.~\ref{fig:discontinuity-management-outliers} we show the detection process starting from the left side of the image, we fit the edge pixels (blue) to a horizontal line (green). When the edge orientation changes from horizontal to vertical, we detect the last edge pixels as outliers (orange) and thus our method detects that we are in a discontinuity. In this case, the red pixel is the last one detected when $|E| > T_{ol}$, where $T_{ol}$ the maximum number of outliers.

Next sub-figure \ref{fig:discontinuity-management-extension1} shows the check done to decide whether we should continue drawing straight or the segment has finished. Purple pixels are the pixels in the discontinuity and are skipped using the Bresenham's algorithm drawing in the current line segment direction. Red pixels are drawn following the edge direction (in green) starting from the first pixel after the discontinuity, $a$. We also mark in red the neighbor pixels used to validate the region using the eigenvalues of $\mM$ if the extension pixels in function \textit{canContinueForward}. Pixels in this synthetic example do not have a uniform gradient direction, thus, the process discards all extensions tested with lengths $J\in [5, 7, 9]$. In the figure we show the last one, $J = 9$. Consequently, the algorithm closes the first segment and continues drawing in the dominant gradient direction upwards. 
When enough pixels are gathered, we fit another segment (Fig.~\ref{fig:discontinuity-management-going-up}). When we reach the top of the image, the segment is extended in the backward direction downwards. When this happens, the mechanism to manage discontinuities is activated again as Fig.~\ref{fig:discontinuity-management-extension2} shows, but this time the region meets all the criteria defined and thus the jump is executed. Fig.~\ref{fig:discontinuity-management-edges} and \ref{fig:discontinuity-management-segments} show respectively the fitted edges and the segments.

One of the limitations of local segment detection approaches is the generation of many small segments produced by gradient discontinuities. The process described above helps us alleviate this.

%This process addresses also helps to skip small occlusions and image noise that can make the image to be broken in several small pieces as is usual in other local approaches.

\subsection{Validation of the generated segment candidates}
\label{sec:method-validation}

\begin{figure}
  \centering
  \includegraphics[width=0.3\columnwidth]{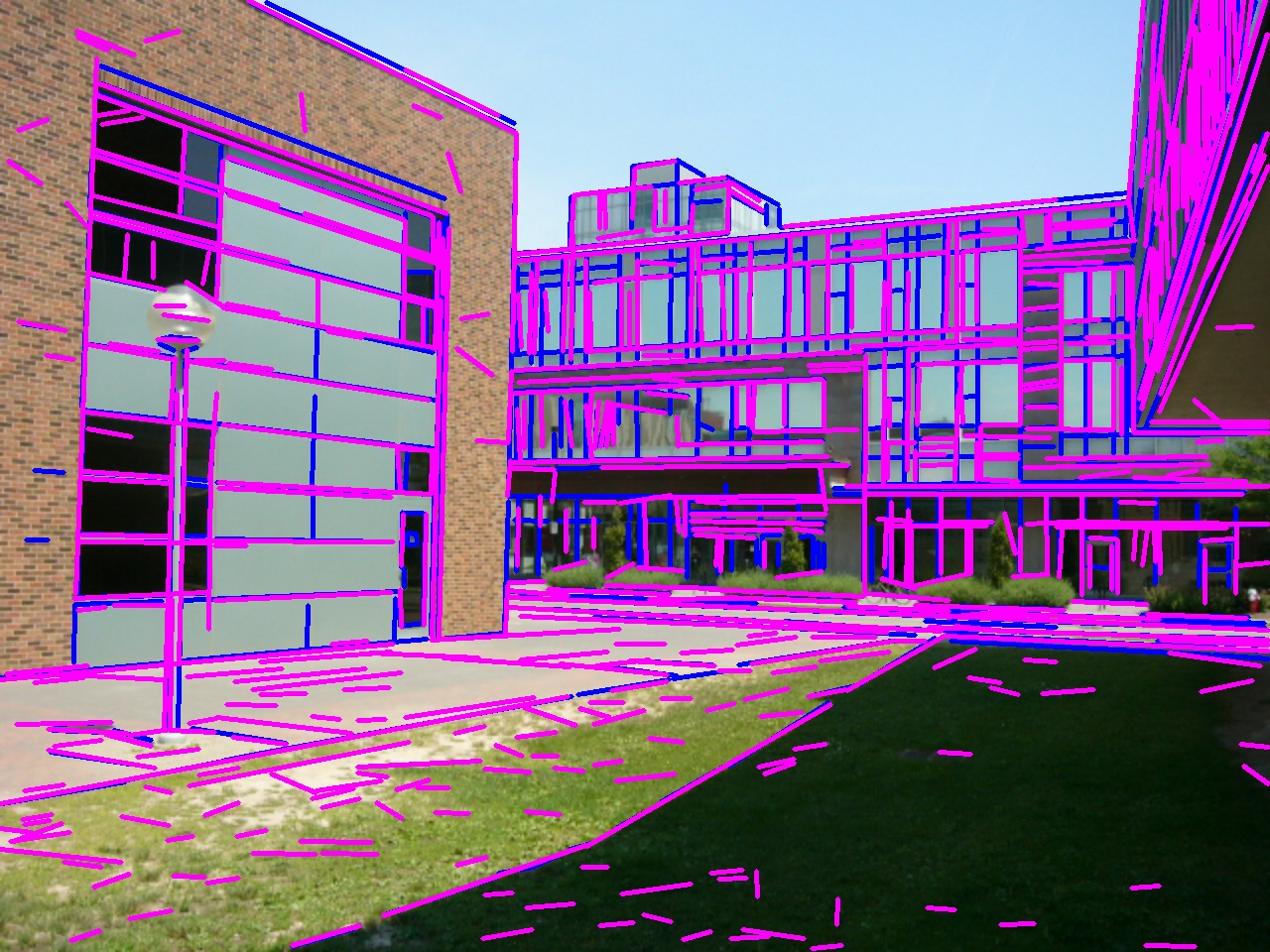}
  \includegraphics[width=0.3\columnwidth]{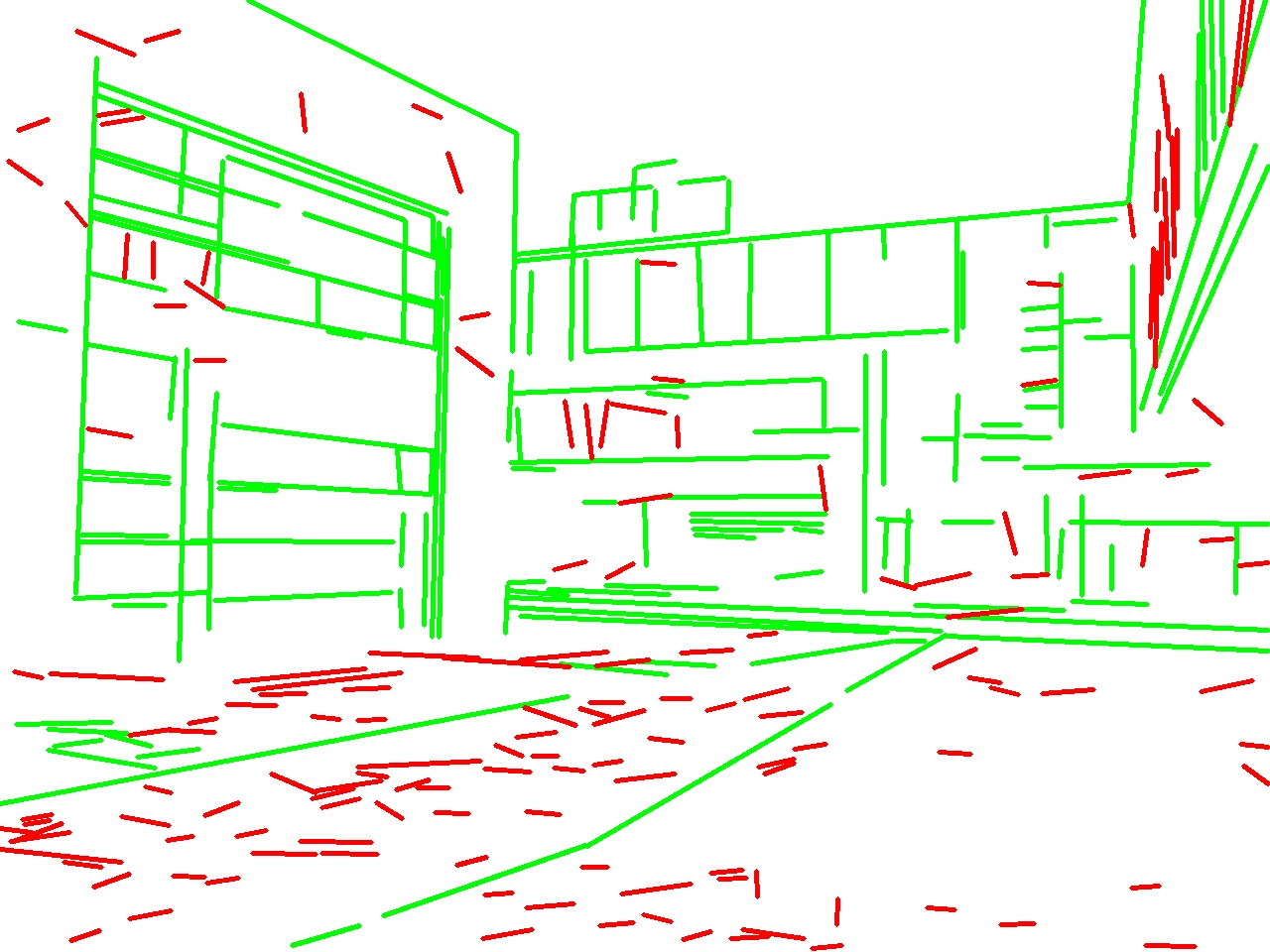}
  \caption{Positive (green) and negative (red) segments detected by ELSED. 
  %The data set is composed of positive (green) and negative (red) segments. 
  We obtain these labels comparing the ground truth segments of the YorkUrban-LineSegment data set (blue) with unvalidated ELSED detections (magenta).}
  \label{fig:positive-and-negative-labels-for-validation}
\end{figure}

After EED, we have several line segments detections, many of them potentially wrong (see red segments in Fig.~\ref{fig:positive-and-negative-labels-for-validation}). They occur mainly in regions with a high density of edge pixels.
To validate a line % several features can be used: gradient magnitude, gradient orientation, intensity variance, gradient variance, color, or a combination of them learnt by a Convolutional Neuronal Network. However, since the main goal of this paper if efficiency, we restrict the measurements to the gradient orientation 
we use the segment pixels' gradient orientation, comparing its direction with the one normal to the segment, i.e. the angular error. This validation can be performed efficiently, without damaging the overall performance.

\begin{figure}
    \centering
    \includegraphics[width=0.45\columnwidth]{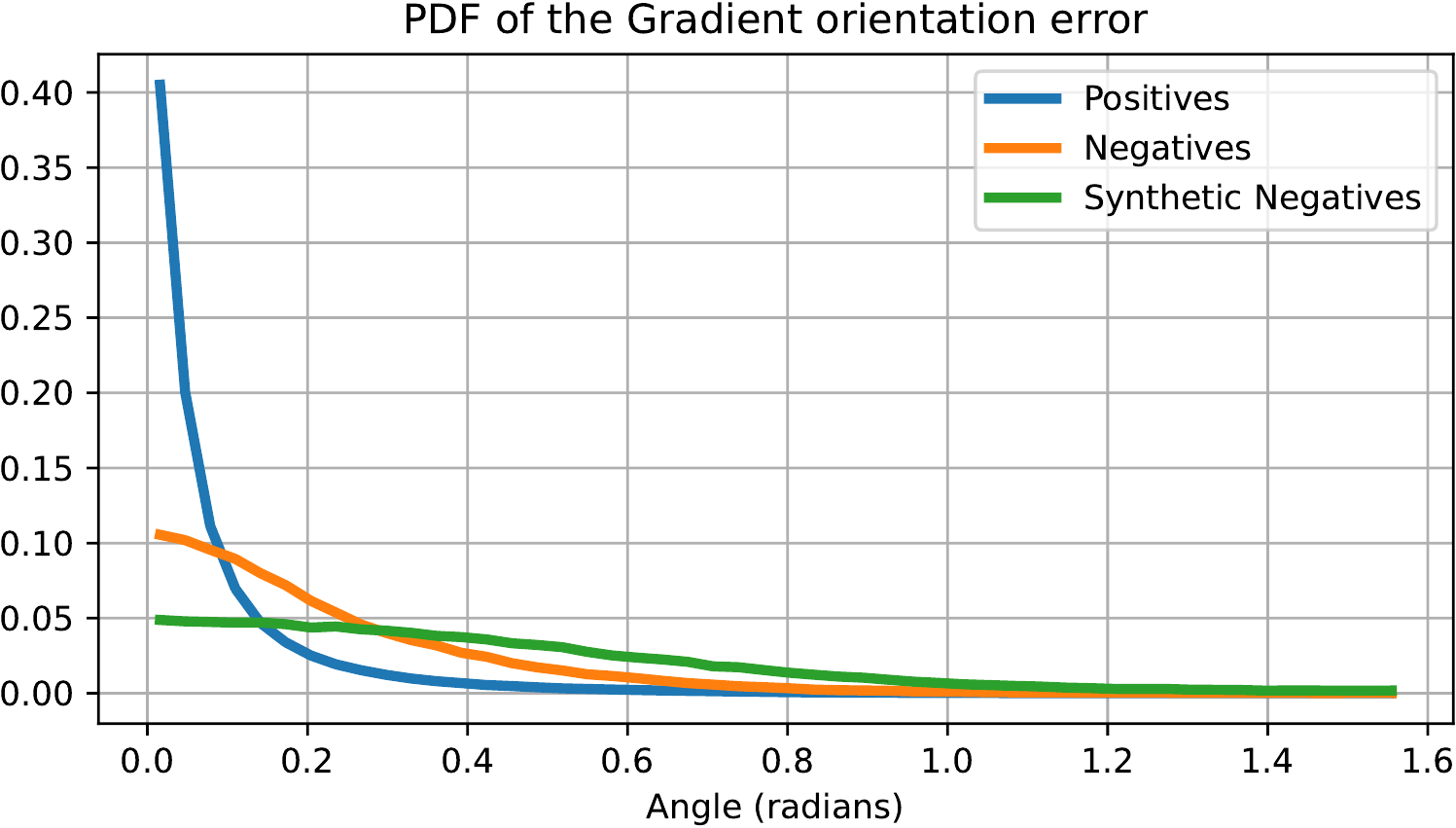}
    \caption{PDF of the gradient orientation error in correct segments (blue), negative segments (orange) and segments detected in a random intensity image (green). }
    \label{fig:gradient-orientation-validation}
\end{figure}

For a good validation we discard the pixels lying in a discontinuity and those near the endpoints, because they usually have a different gradient orientation even in correct detections.
Despite this, the gradient orientation error in a true segment is a noisy signal. 
In Fig.~\ref{fig:gradient-orientation-validation} we can see the probability distribution function (PDF) of orientation errors for pixels on true positive segment detections (TP) (blue), false positive segments (FP) (orange) and false positives segments detected in a random noise intensity image (green). It is clear that the pixels on TP segments have less angular error than those  on FP ones. However, there is a significant overlap between FP and TP distributions. 

Therefore, we use a validation criteria robust to noise. We validate a segment if at least 50\% of its pixels have an angular error lower than a threshold, $T_{valid}$. To separate positive detections (i.e. valid ones) from negative ones, we learn a threshold $T_{valid} = 0.15$ radians which keeps a high recall discarding few true detections.

\subsection{Parameter selection}

Most ELSED parameters have been set empirically and do not need to be changed by the user. We use a Gaussian smoothing filter ($\sigma$ = 1, kernel size = 5$\times$5), a gradient threshold $T_{grad} = 30$, anchor threshold $T_{anchor} = 8$ and $SI$ = 2, that defines the scan interval of anchor every $SI$ row/column. For the line segment fitting:
$T_{ol}=3$, $T_{minLength}=15$, $T_{LineFitErr}=0.2$, $T_{PxToSegDist}=1.5$ and for validation $T_{EigenExt}=10$, 
$T_{AngleExt}=10^{\circ}$, $T_{valid} = 0.15$ radians.

However, other parameters may be tuned by the user to define the type of segments to be detected;
this is the case for the list of jump lengths
%The algorithm should be provided with a list of jump lengths 
that will be tested in the discontinuity management. Since this is directly related to the size of the Gaussian smoothing kernel in the first step ($5 \times 5$ in our implementation) and the size of the gradient convolution kernel ($3 \times 3$ in our case), we define a set of default values, $(5, 7, 9)$, that in the experiments of section~\ref{sec:ablation-study} provide good results. %be the a good compromise when non looking for full lines or long segments. 

\section{Experiments} \label{sec:experiments}

In this section we introduce a methodology to evaluate the accuracy and repeatability of segment detectors and  follow it to compare our detector with the best in the literature. We also 
%This section evaluates empirically the proposed line segment detector and 
present an ablation study to analyze how each component of our algorithm contributes to the final result.

We perform our evaluation in two dimensions, accuracy and efficiency. To this end we have grouped the algorithms in two sets, following the two clusters in Fig.~\ref{fig:sample_of_line_segments}. In the first set we find algorithms that run efficiently in CPU (LSD, EDLine, AG3line and ELSED). In the second set those that require more than one second to process an image (MCMLSD, Linelet, HAWP, SOLD$^2$, F-Clip). 
Although HAWP, SOLD$^2$ and F-Clip are DL methods and should be run in GPU, we run them also on CPU to show the different computational requirements of each approach. It is also important to note, that most low-power devices like smartphones, drones or IoT devices are usually not prepared to run the GPU for long periods of time.
In our experiments we compare the accuracy and efficiency of ELSED with the approaches in each group.
%

%%%%%%%%%%%%%%%%%%%%%%%%%%%%%%%%%%%%%%%%%%%%%%%%%%%%%%%%%%%%%%%%%%%%%%%%%%%%%%
%%%%%%%%%%%%%%%%%%%%%%%%%%%%%%%%%%%%%%%%%%%%%%%%%%%%%%%%%%%%%%%%%%%%%%%%%%%%%%
%%%%%%%%%%%%%%%%%%%%%%%%%%%%%%%%%%%%%%%%%%%%%%%%%%%%%%%%%%%%%%%%%%%%%%%%%%%%%%

%\subsection{\textbf{Meaningful 1-to-1 evaluation of segment detection}}
\subsection{Segment detection evaluation}
\label{sec:exp-seg-detection}

\begin{figure}
  \centering %\medskip
  \begin{subfigure}[t]{0.3\linewidth}
    \centering\includegraphics[width=0.9\linewidth]{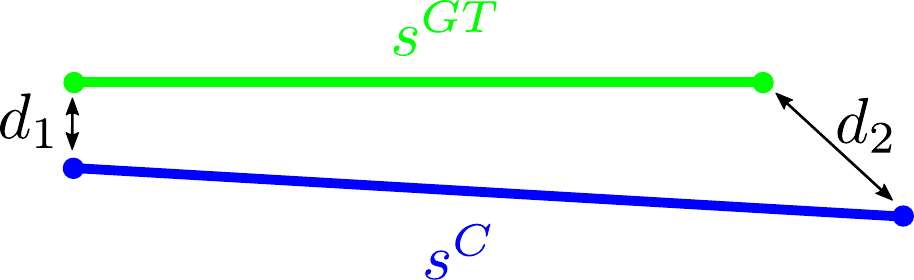}
    \caption{}
%\caption{Structural distance used by ELSED to fill the affinity matrix $\mA$, where each element $\mA(s^C, s^{GT}) = \frac{d_1 + d_2}{2}$.}
    \label{fig:Structural-distance}    
  \end{subfigure}
  \begin{subfigure}[t]{0.3\linewidth}
    \centering\includegraphics[width=0.9\linewidth]{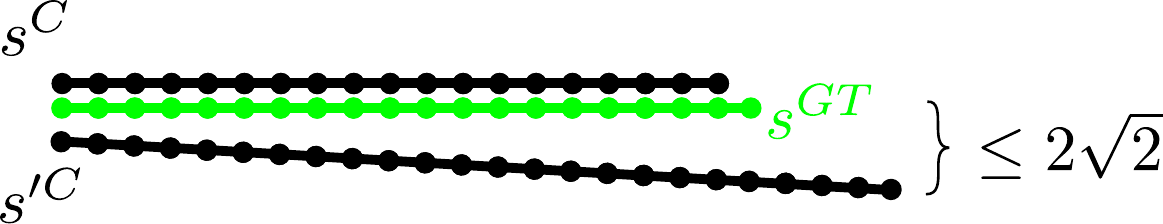}
    \caption{}
  %\caption{Example where MCMLSD~\cite{almazan2017mcmlsd} assignation fails. The detected segment $s^{\prime C}$ is matched to the GT segment $s^{GT}$ while the best possible match is $s^C$. This is because $\mA(s^C, s^{GT}) = 19$ whereas $\mA(s'^C, s^{GT}) = 20$.}
  \label{fig:MCMLSD-countersampple}
  \end{subfigure}
  \label{fig:segments-distance-comparison}
  \caption{Comparison of (a) Structural distance used by ELSED to fill the cost matrix $\mA$ and (b) point-sample based distance used by MCMLSD~\cite{almazan2017mcmlsd}. In (b) the detected segment $s^{\prime C}$ is matched to the ground truth (GT) segment $s^{GT}$ while the best possible match is $s^C$.}
\end{figure}

We evaluate the segment detection in the York Urban Data set (YUD)~\cite{denis2008efficient} that contains indoor and outdoor man-made scenes where some salient segments have been manually labeled. The data set was extensively re-labeled later in Linelet~\cite{cho2017novel}, which contains all segments in the scene, but also some inconsistencies.% such as the labels in tiles and cables, segments smaller than 2 pixels or segments on curves and objects of doubtful value. 

We propose a new evaluation framework that combines the benefits of previous evaluation protocols~\cite{almazan2017mcmlsd, cho2017novel, zhou2019end}, namely, it is fast to compute and is a fair and stable metric for line segment detection.
%
%Our first change is in the way segments are matched. 
We ensure a good 1-to-1 match between detected and ground truth (GT) segments by using the Hungarian algorithm to find the optimal bipartite match. The assignation problem is defined with a cost (or matching) matrix $\mA$, that is filled using the structural distance (see Fig.~\ref{fig:Structural-distance}). This metric is a good trade off between perpendicular distance, misalignment and overlap. It is also faster to compute than the matched number of sampled points.
To speed up the Hungarian algorithm and ensure a meaningful matching, we require matched segments to % be overlapped, have a similar angle and a small perpendicular distance. % but since we consider a 1-to-1 matching, our proximity thresholds can be relaxed. 
%
%\textbf{Esto deberíamos hacerlo? To make a fair comparison, we re-scale the detected line segments to an image of diagonal $128\sqrt{2}$, preserving the aspect ratio.}
%
have an intersection over union bigger than $\lambda_{overlap} = 0.1$, to have an angular distance smaller than $\lambda_{ang} = 15^{\circ}$ and a perpendicular distance smaller than $\lambda_{dist} = 2\sqrt{2}$. The pairs of segments that do not meet these criteria, have infinite cost in the corresponding entry of $\mA$, avoiding their matching. Let $\vx$ be the set of detected segments and $\vy$ the set of ground truth segments. With the 1-to-1 assignations $\mA^*$ between them we define:
\begin{itemize}
    \item \underline{Precision}: Length of the matched intersection measured over the detected segment $\vx_i$, divided by the length of the detected segments, % $|\vx|={\sum_{i}|\vx_i|}$, 
    %\begin{equation}
    $
        \text{P} = \frac{\sum_{i, j \in \mA^* } \vx_i \cap_{\vx_i} \vy_j}{\sum_{i}|\vx_i|}
    $.
    %\end{equation}
    \item \underline{Recall}: Length of the matched intersection measured over the ground truth segment $\vy_j$, divided by the length of the ground truth segments,
    %\begin{equation}
    $
        \text{R} = \frac{\sum_{i, j \in \mA^* } \vy_j \cap_{\vy_j} \vx_i}{\sum_{j}|\vy_j|}
    $.    
    %\end{equation}
    \item \underline{Intersection over Union}: Length of the matched intersection measured over the ground truth segments, divided by length of the matched union measured over the ground truth segments, %
    %\begin{equation}
    $
        \text{IoU} = \frac{\sum_{i, j \in \mA^* } \vy_j \cap_{\vy_j} \vx_i}{ \sum_{i, j \in \mA^* }\vy_j \cup_{\vy_j} \vx_i}
    $.
    %\end{equation}
\end{itemize}

\begin{figure}
    \centering
        \centering
        \includegraphics[width=0.36\textwidth]{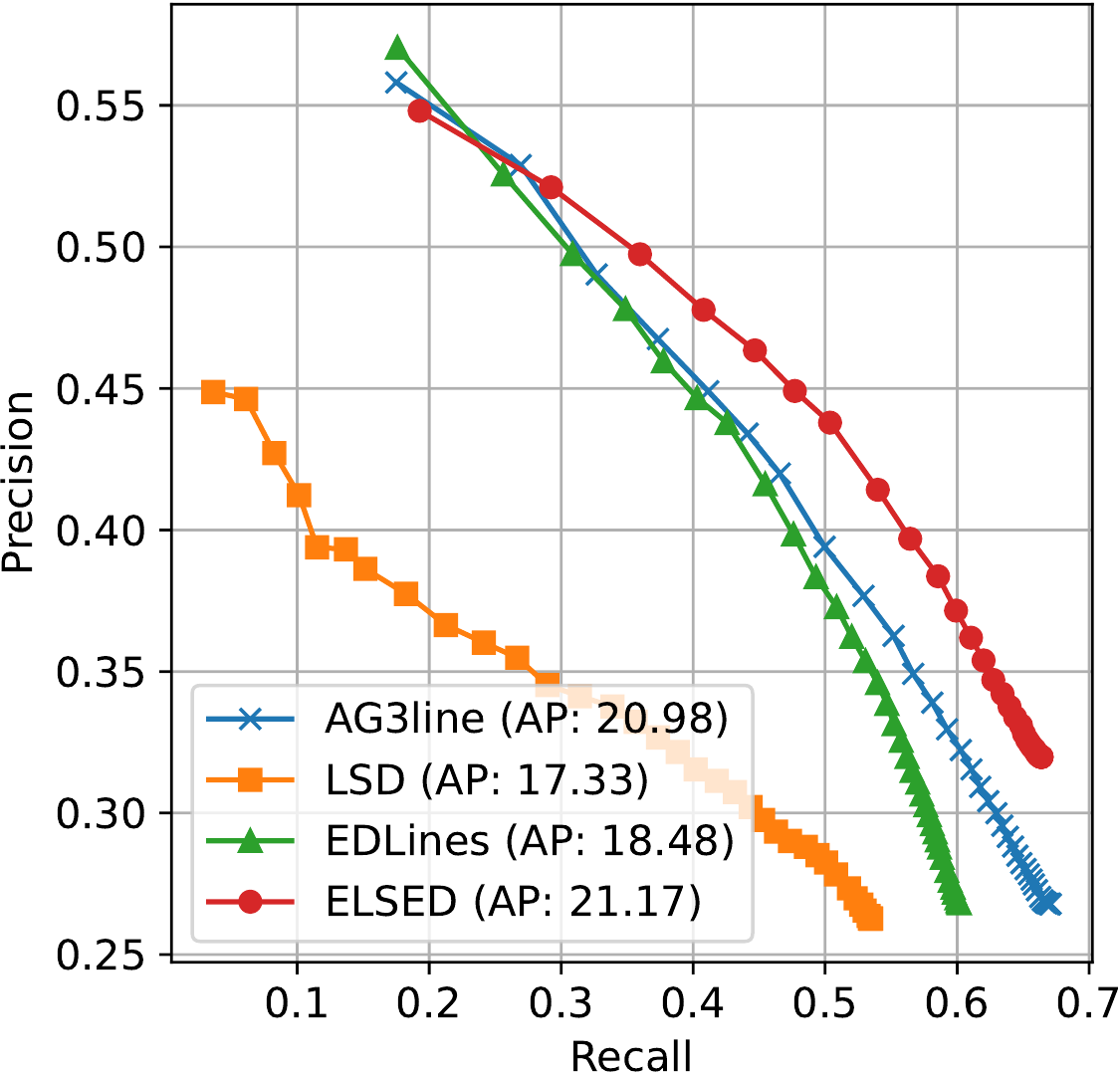}
        \includegraphics[width=0.34\textwidth]{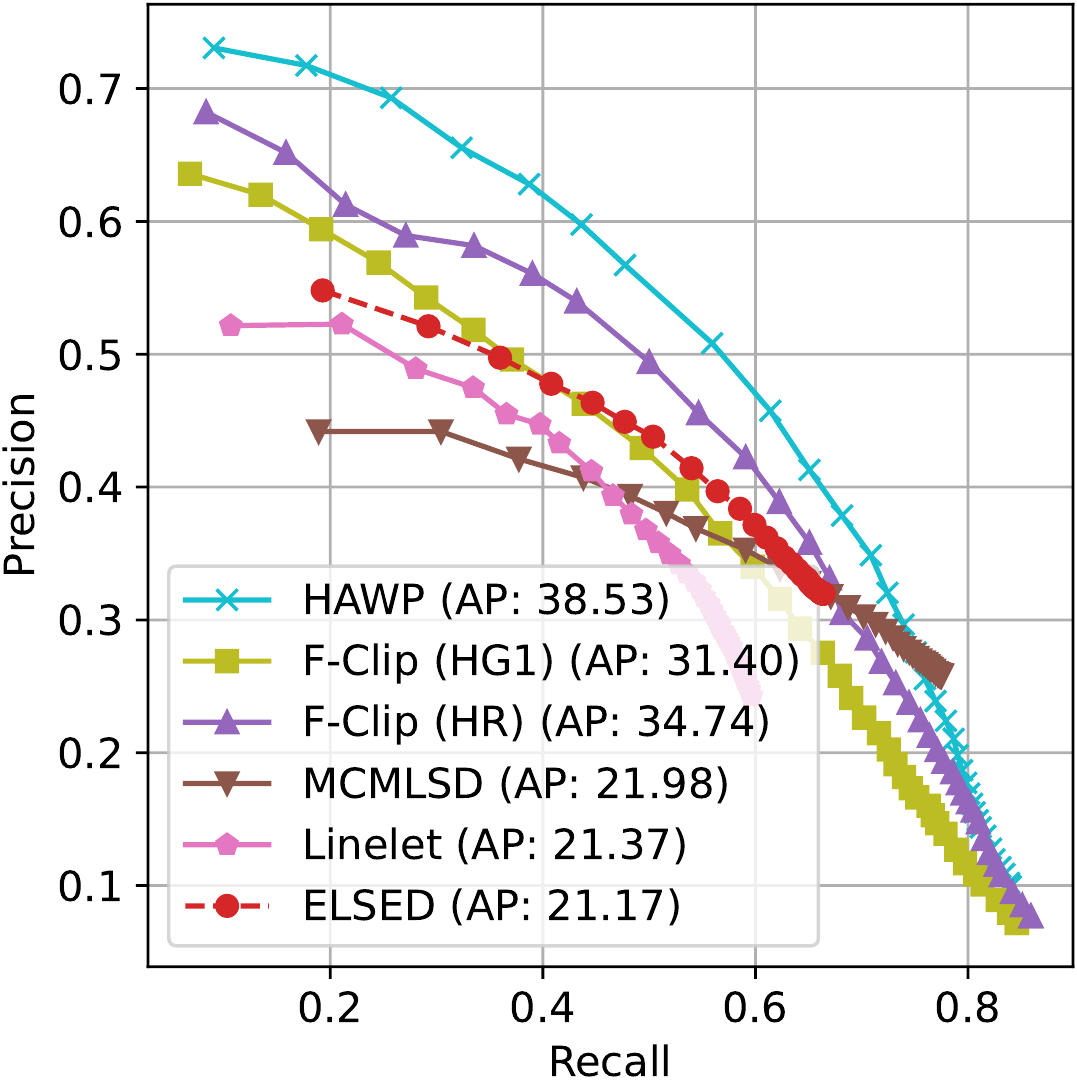}
        \caption{Detected segments' Precision-Recall with our evaluation framework in YUD with original labels.}
        \label{fig:ours-eval-precision-recall}
    % \end{subfigure}    
\end{figure}
In Fig.~\ref{fig:ours-eval-precision-recall} we show the Precision-Recall of our algorithm and the state-of-the-art detectors. This curve is computed w.r.t. the original YUD data set annotations and we sort the segments produced by each method using the score provided by their original code. For ELSED, the score is the percentage of pixels that have an angular error lower than $T_{valid}$.
%\textbf{We use the score of each segment returned by each algorithm (e.g. NFA in the LSD and EDLines, ELSED score is ...)}.
%
%The most important part of the results is the right part where most of the points fall.
%\textbf{Cuesti\'on primordial para este experimento: El AP de las gráficas lo calculo como:} $AP = \sum_{n} (R_n - R_{n-1})P_n$, \textbf{pero esta fórmula solo está bien si el recall empieza en 0 y acaba en 1 porque en ese caso estás haciendo una media ponderada de las precisiones. En nuestro caso esto no es así, por lo que en la comparativa de P-R los métodos que detectan más segmentos y llegan a un mayor recall salen beneficiados en el AP. Os propongo emplear esta otra definici\'on:}  $AP = \frac{\sum_{n} (R_n - R_{n-1})P_n}{ \max_{\forall i}(R_i) - \min_{\forall i}(R_i) }$ que es la que he usado en el Ablation Study.
%
The curve of ELSED is better than those of LSD, EDLines and AG3line, the methods able to run on CPU efficiently (see experiment~\ref{sec:experiment-times}). We obtain a better precision, reaching similar or higher recall. MCMLSD gets better recall but at the cost of very bad precision. This was expected since it is the only method based on the HT and detects all full lines with enough support, even the hallucinated ones over highly textured regions. 
ELSED is able to improve AG3line because it has a better drawing scheme tailored to the line segment detection problem.
ELSED is on par with the most efficient DL approaches, F-Clip (HG1), in the range of recall where ELSED works. Deeper networks such as HAWP that needs a GPU and are far away from real-time in CPU obtain, as expected, results with better recall and precision. 

\begin{table}
\centering
\scriptsize
\setlength{\tabcolsep}{1ex} % JM: Esta es la línea que controla eñ espacio entre columnas
\begin{tabular}{|l|l|l|l|l||l|l||l|l|l|l||l|l|} \hline
& \multicolumn{6}{|c||}{Original YUD annotations} 
& \multicolumn{6}{|c|}{YorkUrban-LineSegment annotations} \\ \hline
Method & P & R & IoU & F\_sc & AP & bAP  & P & R & IoU & F\_sc & AP & bAP \\ \hline
LSD & 0.26 & 0.53 & 0.59 & 0.34 & 17.33 & 34.78 & 0.68 & 0.52 & 0.67 & 0.58 & \textbf{38.23} & 76.35 \\
EDLines & 0.27 & 0.60 & 0.64 & 0.36 & 18.48 & 43.42 & 0.67 & 0.56 & 0.68 & \textbf{0.60} & 36.61 & 76.88 \\
AG3line & 0.27 & \textbf{0.67} & 0.69 & 0.37 & 21.02 & 42.32 & 0.60 & \textbf{0.62} & 0.66 & 0.60 & 34.69 & 69.06 \\
ELSED & \textbf{0.32} & 0.66 & \textbf{0.71} & \textbf{0.41} & \textbf{21.17} & \textbf{44.94} & 0.68 & 0.53 & 0.68 & 0.58 & 31.01 & 71.69 \\ 
ELSED-NJ & - & - & - & - & -  & -  & \textbf{0.71} & 0.51 & \textbf{0.69} & 0.59 & 32.43 & \textbf{76.93} \\ \hline
MCMLSD & 0.26 & 0.77 & 0.74 & 0.37 & 21.98 & 37.50  & 0.55 & \textbf{0.62} & 0.62 & 0.57 & 30.24 & 57.71 \\
Linelet & 0.24 & 0.60 & 0.63 & 0.33 & 21.37 & 43.62 & 0.62 & 0.58 & 0.66 & \textbf{0.59} & \textbf{39.35} & \textbf{75.52} \\
HAWP & \textbf{0.49} & \textbf{0.60} & \textbf{0.83} & \textbf{0.51} & \textbf{38.53} & \textbf{51.41}  & 0.65 & 0.30 & 0.70 & 0.40 & 33.48 & 56.84 \\ 
%$\text{SOLD}^2$ & 0.2301 & 0.3267 & 0.4791 & 0.2515 & 7.72 & 27.30 \\ 
F-Clip(HG1) & 0.47 & 0.42 & 0.80 & 0.42 & 31.40 & 40.35  & 0.67 & 0.22 & \textbf{0.72} & 0.32 & 32.13 & 50.06\\ 
F-Clip(HR) & 0.53 & 0.47 & 0.82 & 0.47 & 34.74 & 44.77  & \textbf{0.69} & 0.22 & 0.72 & 0.33 & 32.78 & 51.15 \\ \hline
\end{tabular}
\caption{Results with our evaluation protocol (top half: efficient methods, bottom half: slow methods).% We use 1-to-1 matching and $\lambda_{overlap}=0.1$. 
We use \textbf{bold} for best results in each experiment. The columns show for each method the Precision (P), Recall (R), Intersection over Union (IoU), F-Score (F\_sc), Average Precision (AP) and bounded Average Precision (bAP).}
\label{table:ours_segment_detection_yud}
\end{table}

In Table~\ref{table:ours_segment_detection_yud} columns 2 to 5 show the highest recall values in the P-R curve for each method.
They correspond to the case where we require the detector to find as many segments as possible. The last two columns show metrics over all the P-R points of the curve. 
If we look at the results with the original YUD labels, ELSED has the best precision (0.3198), F-score (0.4148) and IoU (0.7111) among the efficient detectors on CPU. It also obtains the best results for the overall metrics along all the Precision-Recall points (Average Precision, AP). AP is the usual classification metric, that is biased towards detectors with a wider recall range. Thus, we also show the AP bounded to the recall interval where the curve of each method is defined (bAP). ELSED has a bAP comparable with F-Clip (HR) and better than F-Clip (HG1). This means that, although ELSED detects fewer segments (lower recall range than F-Clip) it has a precision similar to the top DL-based methods in the detected segments.

We also present the results for the ''YorkUrban-LineSegment annotation''~\cite{cho2017novel}. %, that extends York-Urban dataset with short and broken segments. 
In the new annotation, the definition of segments changes to short and broken lines from the global lines in the original YUD labels. In this case, we also show the results of ELSED without jumps (\emph{ELSED-NJ}). %, to avoid ELSED to merge segments labeled as separate ones in the data set. 
ELSED-NJ obtains competitive results, being the best in terms of precision and IoU in this data set. This shows the nice property of ELSED, which allows to adapt the segment definition to the application by changing the jump length over discontinuities.

With these experiments we can conclude that, although ELSED has been designed with the objective of reducing the execution time, it is also a competitive algorithm in terms of segment detection accuracy. The reason is that the EED process is adapted to the segment detection problem and jumps over discontinuities to produce outputs that match the segment length in the annotations.

%%%%%%%%%%%%%%%%%%%%%%%%%%%%%%%%%%%%%%%%%%%%%%%%%%%%%%%%%%%%%%%%%%%%%%%%%%%%%%
%%%%%%%%%%%%%%%%%%%%%%%%%%%%%%%%%%%%%%%%%%%%%%%%%%%%%%%%%%%%%%%%%%%%%%%%%%%%%%
%%%%%%%%%%%%%%%%%%%%%%%%%%%%%%%%%%%%%%%%%%%%%%%%%%%%%%%%%%%%%%%%%%%%%%%%%%%%%%
\subsection{Wireframe parsing}
\label{sec:wireframe}
\begin{figure}
    \centering
    \begin{subfigure}{0.35\linewidth}
    \centering\includegraphics[width=\linewidth]{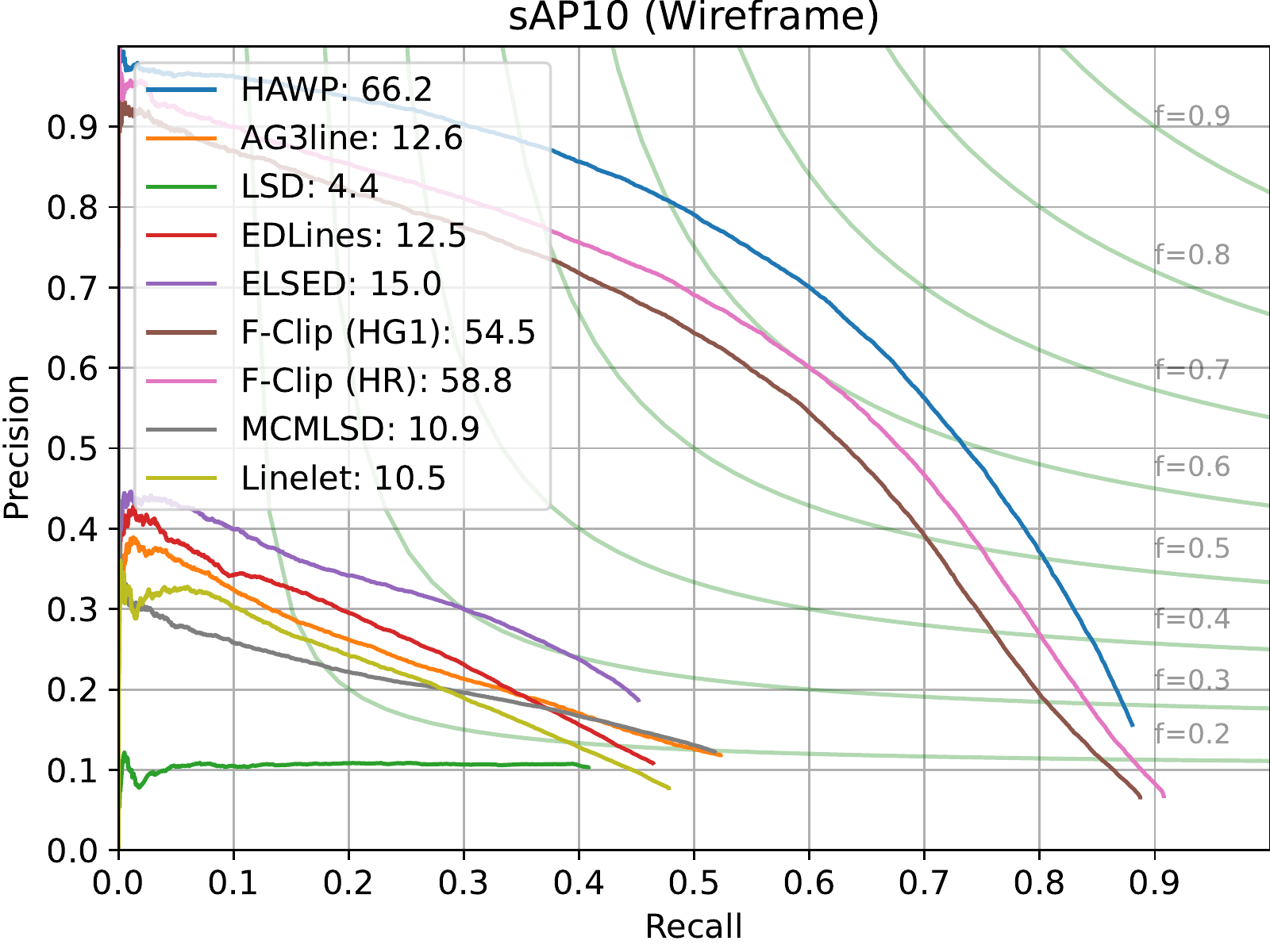}
    \caption{Wireframe}
    \label{fig:sap-eval-wireframe}    
    \end{subfigure}
    \begin{subfigure}{0.35\linewidth}
    \centering\includegraphics[width=\linewidth]{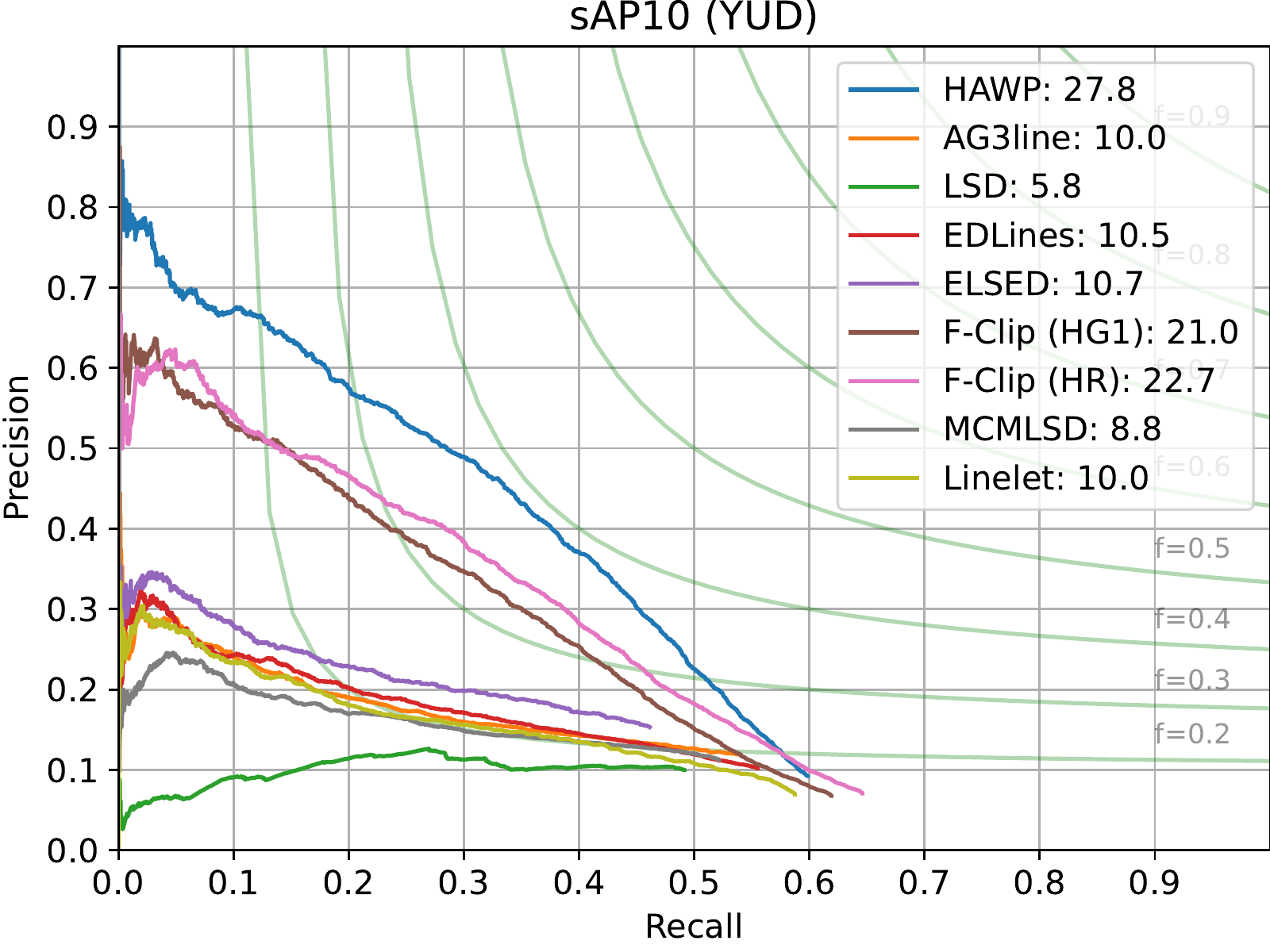}
    \caption{York Urban Dataset}
    \label{fig:sap-eval-yud}
    \end{subfigure}
    \caption{Wireframe parsing evaluation. }
    \label{fig:sap-eval}
\end{figure}

This experiment presents a comparative evaluation in a different task: Wireframe Parsing. %
Figure \ref{fig:sap-eval-wireframe} shows the evaluation results in the ShanghaiTech Wireframe dataset~\cite{huang2018wireframe} and YUD, with the evaluation protocol described in L-CNN~\cite{zhou2019end}. Unlike the protocol presented in section \ref{sec:exp-seg-detection}, here all segments have the same importance and the GT segments are matched to detections by nearest neighbour.

As expected, computationally-expensive methods (HAWP and F-Clip), trained for this task, achieve top-performing results.
However, when the same evaluation protocol is applied in a similar dataset, YUD (Fig. \ref{fig:sap-eval-yud}),  their precision roughly halves. Partly because YUD's annotations are less dense, but also as result of data acquisition and labeling biases.
By contrast, general-purpose line segment detectors generalize better. ELSED obtains the best results among them. This is because EED is able to retrieve more segments from each anchor and because the jump strategies properly manage gradient noise.

%%%%%%%%%%%%%%%%%%%%%%%%%%%%%%%%%%%%%%%%%%%%%%%%%%%%%%%%%%%%%%%%%%%%%%%%%%%%%%
%%%%%%%%%%%%%%%%%%%%%%%%%%%%%%%%%%%%%%%%%%%%%%%%%%%%%%%%%%%%%%%%%%%%%%%%%%%%%%
%%%%%%%%%%%%%%%%%%%%%%%%%%%%%%%%%%%%%%%%%%%%%%%%%%%%%%%%%%%%%%%%%%%%%%%%%%%%%%

\subsection{Repeatability}
\label{sec:repeatability}

Regardless the type of segments aimed for the detection, a desirable property is the robustness against changes in viewpoint, scale, rotation or lighting. In this subsection we evaluate the detectors' repeatability. Given two images of the same scene under different conditions, the capacity to detect the same segments in both situations. Specifically, given two images, we define line segments repeatability as the ratio between the length of 1-to-1 segment matches and the total length of segments detected in both images. We take into account only the segments located in the part of the scene present in both images, adjusting their endpoints accordingly. 

We use the images of HPatches~\cite{hpatches2017cvpr} %data set that mainly contains human-made environments with viewpoint and illumination changes from 116 sequences, 6 images per sequence. We match the first image of each sequence with the other 5 obtaining a repeatability value for each. The
where the repeatability of segment detections in images $\mathcal{A}$ and $\mathcal{B}$ is computed as:
\begin{equation}
    \text{repeatability} = \frac{ \sum_{i, j \in \mA^* }
    \vx_i^{\mathcal{A}} \cap_{\vx_i^{\mathcal{A}}} \vx_j^{\mathcal{A} | \mathcal{B}}  }{ \sum_i \left|\vx_i^{\mathcal{A}}\right| + \sum_j \left|\vx_j^{\mathcal{A} | \mathcal{B}}\right| } +
    \frac{\sum_{i, j \in \mB^* } \vx_i^{\mathcal{B}} \cap_{\vx_i^{\mathcal{B}}} \vx_j^{\mathcal{B} | \mathcal{A}}  }{ \sum_i \left|\vx_i^{\mathcal{B}}\right| + \sum_j \left|\vx_j^{\mathcal{B} | \mathcal{A}}\right| }
\end{equation}
Where $\vx^{\mathcal{I}}$ are the segments detected in image $\mathcal{I}$ and $\vx^{\mathcal{I} | \mathcal{J}}$ the segments detected in image $\mathcal{J}$, projected to $\mathcal{I}$ using the homography between them. % Note that when we project the segments, we delete those falling outside the image and for the ones in the border we adjust the endpoint out of the image to the border.
The matching matrix $\mA^*$ contains the 1-to-1 segment assignations between segments   $\vx^{\mathcal{A}}$ and $\vx^{\mathcal{A} | \mathcal{B}}$ obtained with the matching process described in subsection \ref{sec:exp-seg-detection} and $\mB^*$ the one between $\vx^{\mathcal{B}}$ and $\vx^{\mathcal{B} | \mathcal{A}}$.

\begin{table}
    \centering
    \scriptsize
    \begin{tabular}{|c|c|c|} \hline
    Method  & Length Repeatability & Num Segs. Repeatability \\ \hline
    LSD             & 0.53934 & 0.48707 \\ 
    EDLines         & 0.54352 & 0.47242 \\
    AG3line         & 0.43582 & 0.40002 \\
    ELSED           & \textbf{0.55928} & \textbf{0.50285} \\ \hline
    MCMLSD          & 0.50322 &  0.39440 \\
    Linelet         & 0.52102 &  0.43427 \\
    HAWP            & 0.40056 & 0.41248 \\ 
    $\text{SOLD}^2$ & 0.46161 & 0.47957 \\ 
    F-Clip (HG1)    & 0.39784 & 0.43560 \\ 
    F-Clip (HR)     & 0.40969 & 0.43413 \\ \hline
    \end{tabular}
    \caption{Mean repeatability of each segment detector in the HPatches data set. The higher the repeatability, the more robust the detector.}% is before changes in lighting, perspective, rotation and scale. }
    \label{tab:repeatability}
\end{table}
 
In Table~\ref{tab:repeatability} the second column also shows the repeatability in terms of the number of matched segments. We employ here $\lambda_{overlap} = 0.5$ and $\lambda_{dist} = 5$ according to \cite{pautrat2021sold2}. It both cases ELSED obtains the most repeatable results. This is because EED provides stability to the edge detection and also because the jump strategy is able to overcome small discontinuities that cause other local methods to fail.
We have also observed that deep models like $\text{SOLD}^2$ %~\cite{pautrat2021sold2}
or HAWP get highly repeatable results in some scenes and very bad results in others, this is possible because they have been trained in a quite specific problem (the wireframe parsing for indoor scenes) and do not generalize well to the diverse images of Hpatches.

%%%%%%%%%%%%%%%%%%%%%%%%%%%%%%%%%%%%%%%%%%%%%%%%%%%%%%%%%%%%%%%%%%%%%%%%%%%%%%
%%%%%%%%%%%%%%%%%%%%%%%%%%%%%%%%%%%%%%%%%%%%%%%%%%%%%%%%%%%%%%%%%%%%%%%%%%%%%%
%%%%%%%%%%%%%%%%%%%%%%%%%%%%%%%%%%%%%%%%%%%%%%%%%%%%%%%%%%%%%%%%%%%%%%%%%%%%%%

\subsection{Ablation study}
\label{sec:ablation-study}

%In this section we present an ablation analysis of the components involved in our algorithm (see Table~\ref{table:ablation_study}). 
%
%
\begin{table}[t]
\centering
\scriptsize
\resizebox{\linewidth}{!}{%
%\rowcolors{1}{white}{mygray}
\begin{tabular}{|c|c|c||c|c|c|c||c|c|} \hline
\multicolumn{3}{|c|}{\textbf{Configuration}} & \multicolumn{4}{c|}{\textbf{Last P-R point metrics}} & \multicolumn{2}{c|}{\textbf{Global metrics}} \\ \hline
Jumps & Jump Val. & Seg. Val. & P & R & IoU & F\_sc & bAP & Time (ms)\\ \hline \hline
%\multicolumn{9}{|c|}{Original YUD Labels} \\ \hline
 None     &            &            & 0.271 & 0.609 & 0.646 & 0.360 & 43.06 & \textbf{2.95} \\ \hline
 None     &            & \checkmark & 0.307 & 0.590 & 0.651 & 0.388 & 44.27 & 4.16 \\ \hline
 Fixed (5 px)    &            &            & 0.278 & \textbf{0.702} & \textbf{0.712} & 0.383 & 42.46 & 3.25 \\ \hline
 Fixed (5 px)   & \checkmark &            & 0.282 & 0.667 & 0.694 & 0.381 & 43.62 & 3.48 \\ \hline
 Multiple (5,7,9 px)& \checkmark &            & 0.285 & 0.686 & 0.706 & 0.387 & 43.28 & 4.16 \\ \hline
 Multiple (5,7,9 px) & \checkmark & \checkmark & \textbf{0.320} & 0.664 & 0.711 & \textbf{0.415} & \textbf{44.94} & 5.38 \\ \hline
\end{tabular}
}
\caption{Results of the ablation study using our 1-to-1 evaluation protocol 
over the original YUD % York Urban Data set % and the YorkUrban-LineSegment Labels 
annotations. Best results in 
\textbf{bold}. Execution times measured on Intel Core i7. % \textbf{TODO: Os propongo borrar la parte de YorkUrban-LineSegments Labels.}
}
\label{table:ablation_study}
\end{table}

We start with the simplest version of ELSED: no discontinuity jumps and no validation step (see Table~\ref{table:ablation_study}). This version of the algorithm obtains the worst results (IoU$=0.646$ and F\_sc$=0.360$ in Table~\ref{table:ablation_study}) for long segment detections. % in the original YUD annotations. %However, it the best one (IoU$=0.413$ and F\_sc$=0.587$ in Table~\ref{table:ablation_study}) in the YorkUrban-LineSegment annotations where the segments are typically short ones. 
%
%The YUD dataset annotations are long segments. Thus, not having the capability of deal with discontinuities gives the worst IoU (0.224) and F\_sc (0.373) on that dataset. 
When the validation is activated, the precision increases from 0.271 to 0.307 whereas the recall remains high (0.59). 

\begin{figure*}
    \centering
    \includegraphics[width=0.19\textwidth]{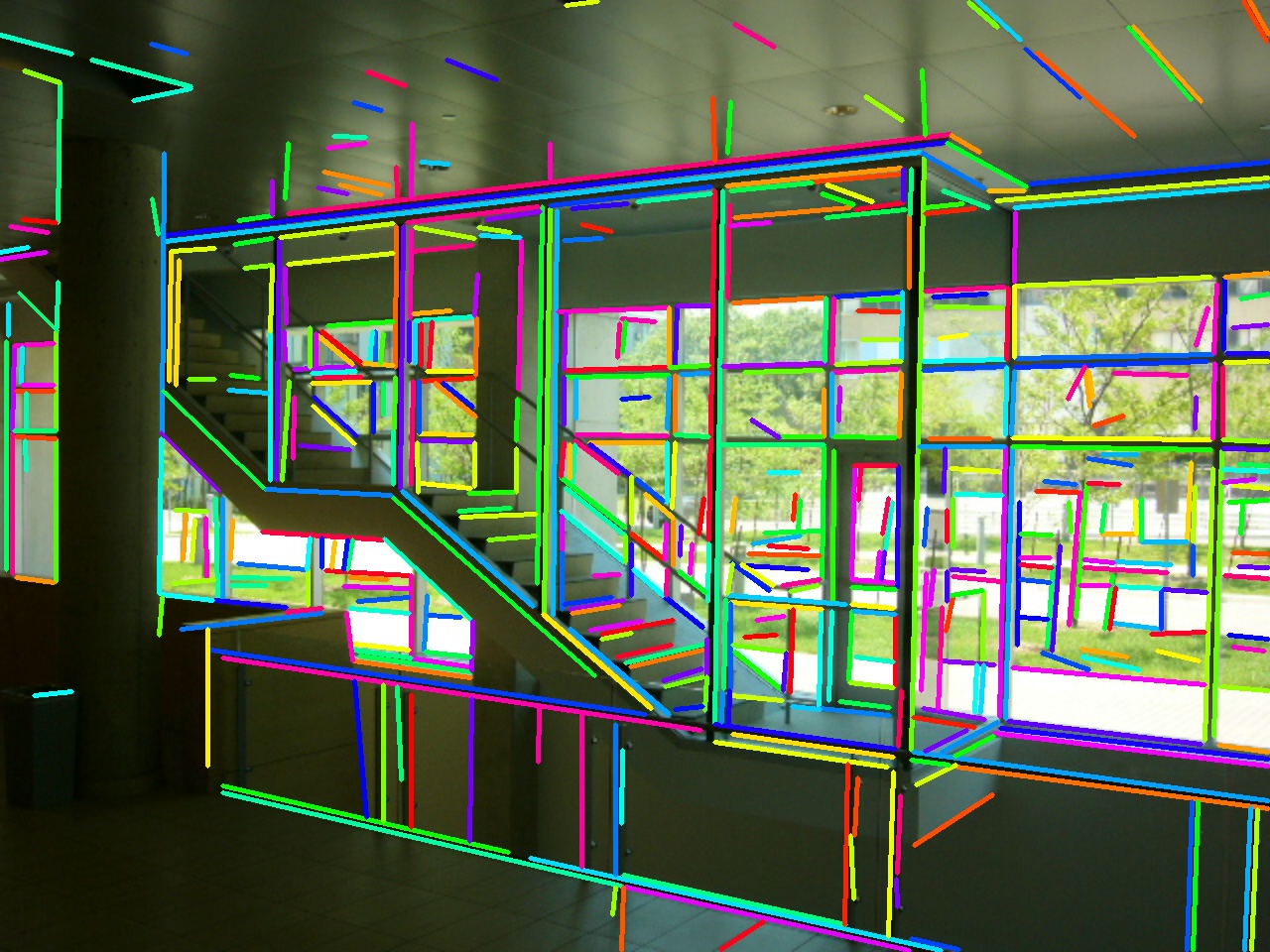}
    \includegraphics[width=0.19\textwidth]{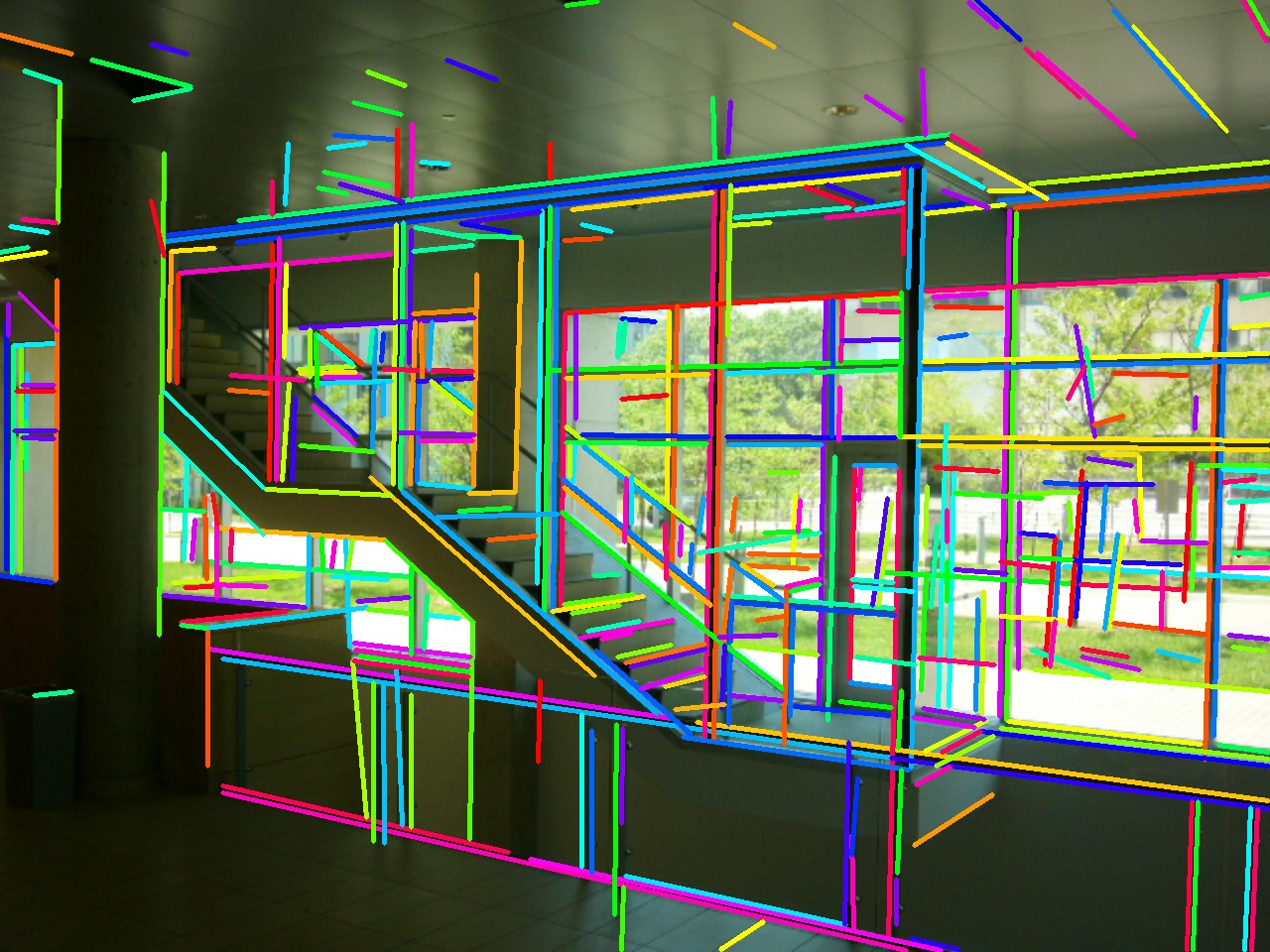}
    \includegraphics[width=0.19\textwidth]{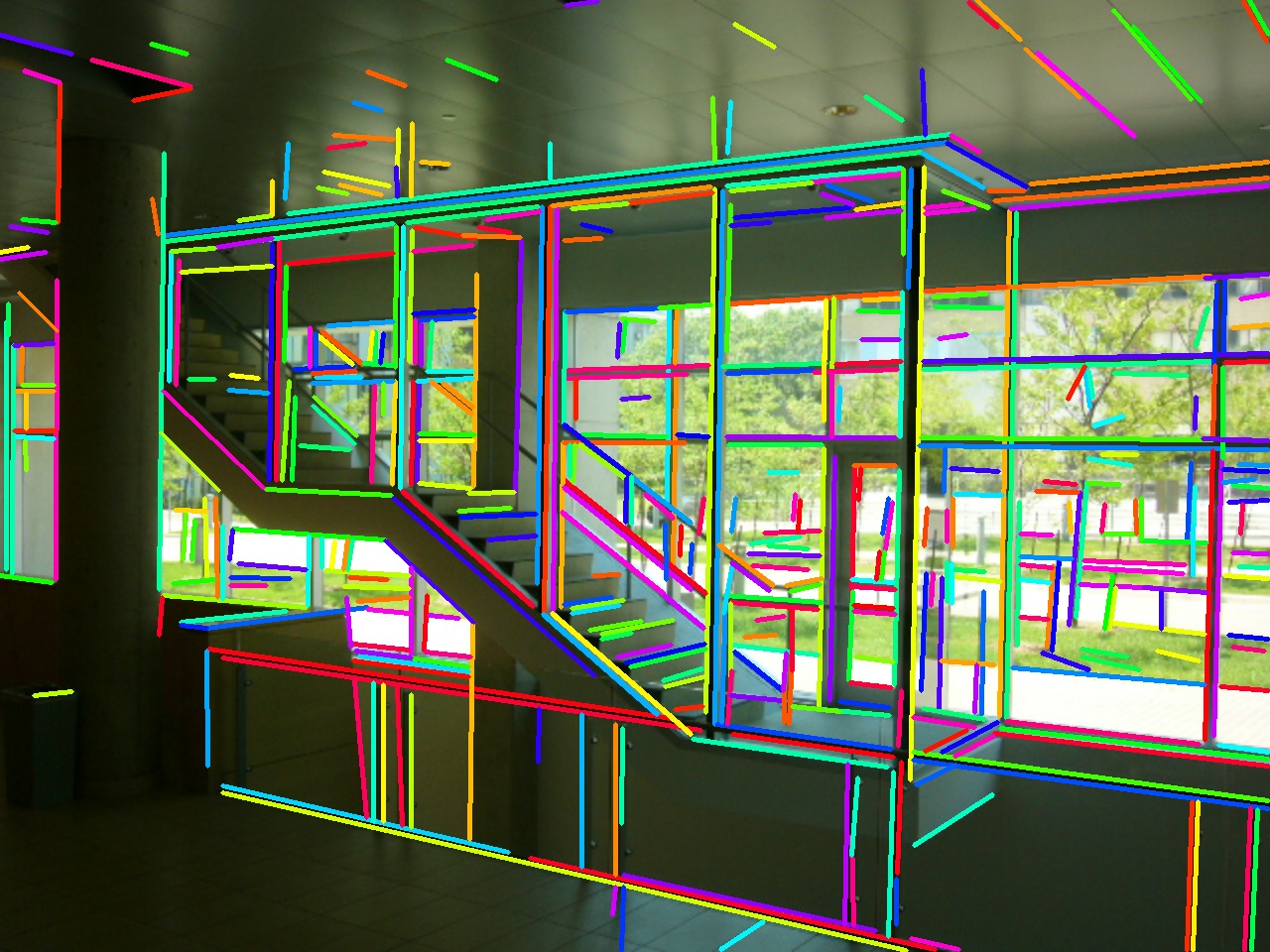}
    \includegraphics[width=0.19\textwidth]{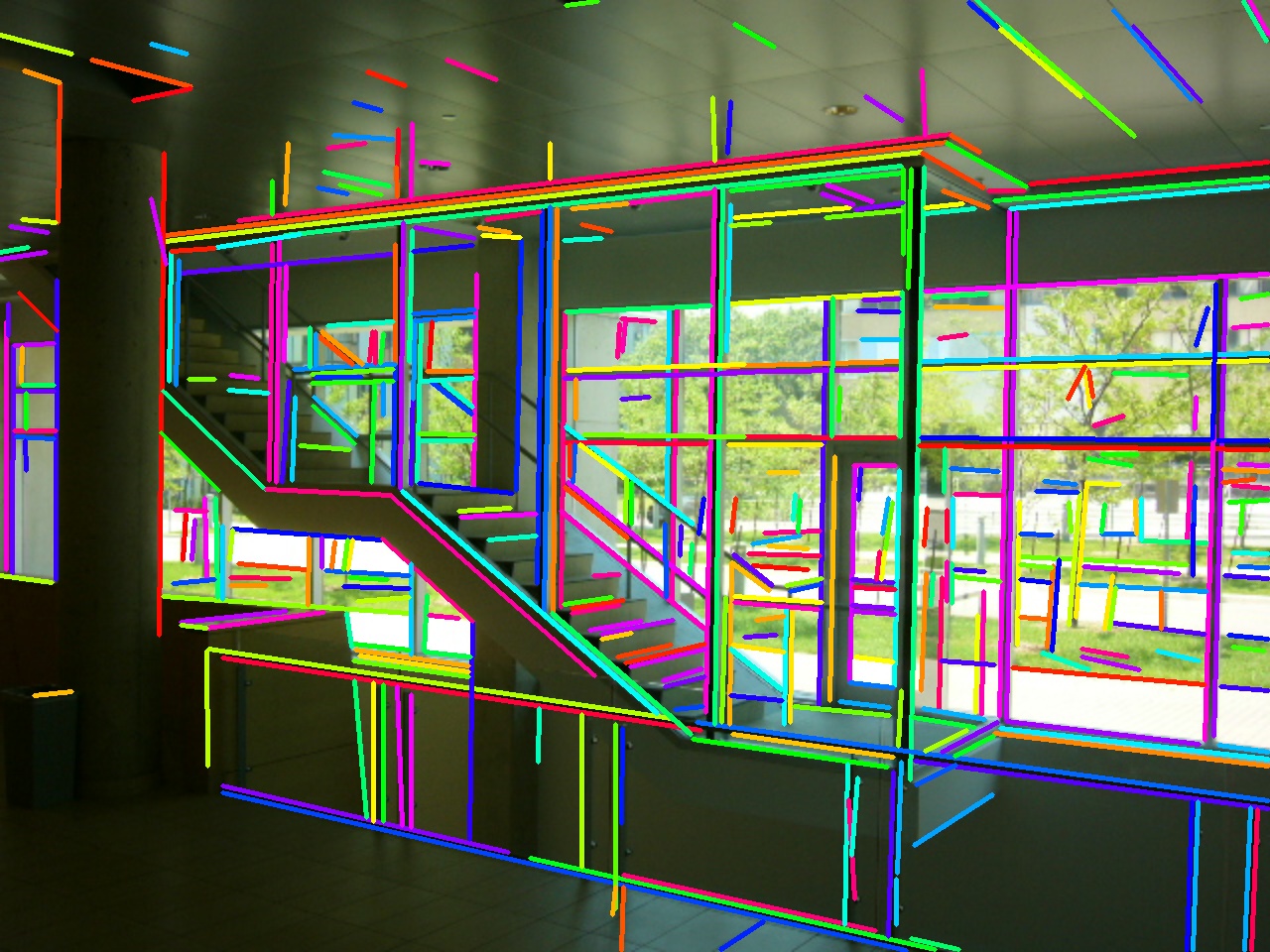}
    \includegraphics[width=0.19\textwidth]{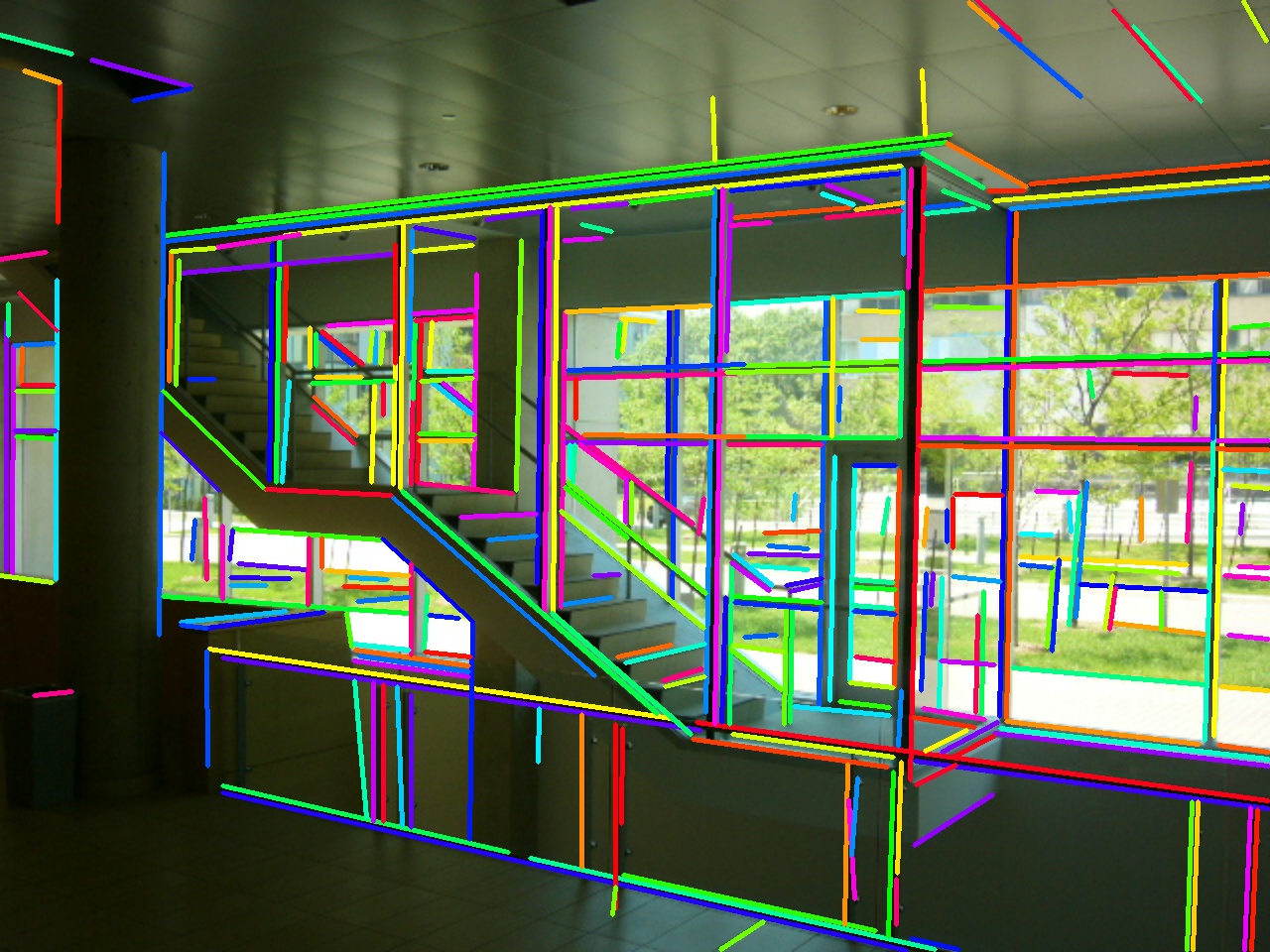}
    \caption{From left to right: No Validation no jumps, No validation with fixed size jumps (5 px) and no jump validation, No validation with fixed size jumps (5 px) and jump validation, No validation with multi-size jumps (5, 7, 9 px) and jump validation, Validation with multi-size jumps (5, 7, 9 px) and jump validation}
    \label{fig:images_ablation_study}
\end{figure*}

If we now add the discontinuity jump component with a fixed length of 5 pixels it removes some small detection errors. For example, in the second row and column of Fig.~\ref{fig:images_ablation_study} the broken segments of the wall in the left, with the added fixed-length jump capability are detected as a unique segment. On the other hand, now the algorithm performs some incorrect jumps going beyond the endpoint of the segment. This effect can be observed in the results of Table~\ref{table:ablation_study} where the Recall takes a big leap (from 0.609 to 0.702) and the Precision also increases moderately (from 0.271 to 0.278). To fix the problem with incorrect jumps, we add the validation of the jump destination region of section~\ref{sec:discontinuity_management} (see the well-fitted endpoints of the third column in Fig.~\ref{fig:images_ablation_study}).  %\textbf{JM: ¿Donde están esos endpoints? Hay que explicarlo mejor}.

On the other hand, a jump of 5 pixels is not big enough if the discontinuity is large. In this case, the jump validation is performed over pixels on the discontinuity. Thus, since discontinuities contain gradients in different directions to the normal of the segment, the jump validation fails. This is the reason why we add the multi-length jumps in the fourth column in Fig.~\ref{fig:images_ablation_study}. With it, we can deal with longer segment discontinuities. The last step is the validation of the whole detected segment (see section~\ref{sec:method-validation}) shown in the last column of Fig.~\ref{fig:images_ablation_study}. 
%In Table~\ref{table:ablation_study} 
Segment validation increases the precision (from 0.285 to 0.320) with a small penalty in the recall (from 0.686 to 0.664).

%%%%%%%%%%%%%%%%%%%%%%%%%%%%%%%%%%%%%%%%%%%%%%%%%%%%%%%%%%%%%%%%%%%%%%%%%%%%%%
%%%%%%%%%%%%%%%%%%%%%%%%%%%%%%%%%%%%%%%%%%%%%%%%%%%%%%%%%%%%%%%%%%%%%%%%%%%%%%
%%%%%%%%%%%%%%%%%%%%%%%%%%%%%%%%%%%%%%%%%%%%%%%%%%%%%%%%%%%%%%%%%%%%%%%%%%%%%%

\subsection{Efficiency evaluation}
\label{sec:experiment-times}

Nowadays CV applications not only require good accuracy, but also fast execution times and low energy consumption. %This is why some methods like LSD~\cite{grompone2010lsd} and EDLines~\cite{akinlar2011edlines} are so popular. This experiment compares the execution times of the most important segment detectors when running on hardware with different computational and energy performance.
This experiment measures the average execution time in the images of YUD data set (Table \ref{table:segment_detection_times}) %for different detectors in different platforms, which is a realistic way to evaluate the computational complexity of an algorithm.
%: 1) the C++ implementation of our detector, 2) the LSD C++ implementation from the authors\footnote{\textbf{LSD Implementation:} \url{https://www.ipol.im/pub/art/2012/gjmr-lsd}} and 3) the original EDLines implementation also from their authors\footnote{\textbf{EDLines implementation}: \url{https://github.com/CihanTopal/ED_Lib}}. 
%YUD 
which contains 101 images with a resolution of $640 \times 480$ pixels. Execution time is measured on four different platforms: a laptop with an Intel Core i7 8750H CPU, 12 cores and 16GB of RAM; a smartphone \textit{Samsung J5 2017} with an Exynox Octa S CPU, 8 cores and 2GB of RAM; a smartphone \textit{One Plus 7 Pro} with Snapdragon 855 CPU, 8 cores and 6GB of RAM and a GPU GeForce GTX 1050 with 4GB of RAM. 

\begin{table}
\centering
\scriptsize
\begin{tabular}{|l|c|c|c|}
\hline
\textbf{Method} & \textbf{Intel Core i7} & \textbf{Snapdragon} & \textbf{Exynox} \\
\hline
LSD         & 36.51	($\pm$1.60)   & 58.68  ($\pm$0.81) &  390.91 ($\pm$0.92) \\
EDLines     & 7.64 ($\pm$0.33)    & 13.79  ($\pm$0.15) &  65.79 ($\pm$0.16)  \\ 
AG3line     & 13.04 ($\pm$0.76)   & 18.57  ($\pm$0.20) &  100.54 ($\pm$0.19) \\ 
%EDLines UPM & 3.58 ($\pm$0.16)    & 7.76  ($\pm$0.04)  &  36.01 ($\pm$0.10)  \\ 
ELSED-NJ & \textbf{4.18 ($\pm$0.23)} & \textbf{8.28 ($\pm$0.03)}  &  \textbf{45.84 ($\pm$0.02)} \\ 
%ELSED-Th20  & 6.19 ($\pm$0.36)    & 11.04  ($\pm$0.08) &  67.39 ($\pm$0.12)  \\ 
ELSED       & 5.38 ($\pm$0.30)    & 10.20  ($\pm$0.07) &  59.99 ($\pm$0.16)  \\ \hline %\cline{3-4}
MCMLSD     & 4.68K ($\pm$1.78K) & \multicolumn{2}{c|}{\textbf{GeForce GTX 1050}} \\ 
Linelet     & 20.9K ($\pm$10.1K) & \multicolumn{2}{c|}{} \\ \cline{3-4}
HAWP        & 12.4K ($\pm$0.8K) & \multicolumn{2}{c|}{212.25 ($\pm$8.35) } \\
$\text{SOLD}^2$ & 3.17K ($\pm$0.52K) & \multicolumn{2}{c|}{417.72 ($\pm$6.78) } \\
F-Clip HR &  7.94K ($\pm$0.15K) & \multicolumn{2}{c|}{47.39 ($\pm$1.46) } \\
F-Clip HG1 & 6.78K ($\pm$0.22K) & \multicolumn{2}{c|}{11.00 ($\pm$0.47) } \\ \hline 
% & \multicolumn{3}{c|}{\textbf{GeForce GTX 1050}} \\ \hline
%HAWP            & \multicolumn{3}{c|}{212.25 ($\pm$8.35) } \\
%$\text{SOLD}^2$ & \multicolumn{3}{c|}{417.72 ($\pm$6.78) } \\
%F-Clip HR & \multicolumn{3}{c|}{47.39 ($\pm$1.46) } \\
%F-Clip HG1 & \multicolumn{3}{c|}{11.00 ($\pm$0.47) } \\
%\hline
\end{tabular}
\caption{Executions times for different state of art line segment detectors on different processors. Results are the average processing time per image, in the YUD~\cite{denis2008efficient} images with size $640 \times 480$. }
\label{table:segment_detection_times}
\end{table}

We use the implementation provided by the authors of each method: LSD, EDLines AG3line and ELSED in C++, Linelet and MCMLSD in Matlab and HAWP, $\text{SOLD}^2$ and F-Clip in Python. 
ELSED is implemented in C++ with Python bindings. To keep fast execution times we compute only L1 gradient norm, which is faster than L2, and predominant gradient direction (vertical or horizontal). 
In EED  we fit the segments with a least squares approach oriented vertically or horizontally that we compute incrementally and, if possible, we reuse the top element from $\mathcal{D}_{stack}$ to avoid memory reallocation. %This improves the detection times avoiding the resize of $\mathcal{D}_{stack}$ and also, because we are continuously using the same region of memory, we benefit from the usage of the cache memory. 

In all platforms, ELSED is around 2$\times$ faster than AG3line,  and EDLines, $6\times$ faster than LSD and much faster than MCMLSD and Linelet. DL methods are designed to run in the GPU,
%where some extremely optimized networks like F-Clip HG1 achieve real-time processing, 
however GPU may not always be available in some platforms like drones, IoT or mobile phones and when it is, it usually involves unaffordable energy consumption. % As a proxy of energy consumption we run all methods in the same hardware, an Intel Core i7 8750H CPU. 
%We can see now why we have separated detectors in two groups: the efficient ones (LSD, EDLines, AG3line and ELSED) and the not efficient ones (Linelet, MCMLSD and the DL ones). 
Looking at the CPU times, the DL methods need between 2300$\times$ (HAWP) and 1200$\times$ (F-Clip HG1) more computation than ELSED. Moreover, even when we run the DL methods on a laptop GPU (Geforce GTX 1050) ELSED is still faster than any of the methods. 
Therefore, for limited platforms, ELSED represents the best segment detector, as it is not only faster, but also detects better than the other efficient methods (see Table \ref{table:ours_segment_detection_yud}) obtaining the most repeatable group of segments (Table \ref{tab:repeatability}).

\section{Conclusions}
\label{sec:conclusion}

In this paper we have introduced ELSED, a general-purpose, fast and flexible line segment detector. It processes a 640x480 image in less than 6 ms on a regular PC and around 10 ms on a modern smartphone.
This efficiency arises as a result of joining the processes of edge drawing and segment detection in one single step, with an Enhance Edge Drawing (EED) algorithm conceived for the problem of line segment detection. 

ELSED also includes a scheme to jump over discontinuities. This endows our method with a flexible strategy to cope with different segment length requirements and improves its robustness against occlusions, shadows and glitches, which make all efficient methods to break down. This is important, for example, in a problem such as Vanishing Point estimation, where long and accurate segments are required. 

In segment-based reconstruction, repeatability is a key feature desired in detectors. We have also introduced a repeatability metric and experimentally shown that ELSED is the top performer. 

%Another important problem that ELSED solves for the efficient methods, is the robustness against small image noise and occlusions that cause most of the detectors to break the segments. This is especially important in problems such as Vanishing Points estimation or segment-based reconstruction where the duplicity of geometric information can make the next steps of the algorithm fail.

%ELSED uses very few computational resources, this is a disadvantage when compared with models that are able to achieve better results with a deeper understanding of the scene. However, when it comes to real applications in low-power devices, these methods are not feasible and have poor generalization capabilities. 

Overall, ELSED is the fastest and most repeatable segment detector in the literature. It is however less accurate than other DL-based competitors, which are computationally orders of magnitude less efficient. Yet, since it is a general purpose detector, %not based on machine learning,
it exhibits good  performance on different data sets,
% This efficiency entails a weakness; it is less powerful than other DL approaches. However, since it is general-purpose, it has good generalization capabilities that learned methods do not have. It obtains quite competitive results in segment detection benchmarks, 
achieving an AP at the same level as other algorithms orders of magnitude slower. 
%
%Overall, ELSED is the fastest segment detector in the literature, getting also quite competitive results in segment detection and repeatability benchmarks, reaching a state of art accuracy at the same level as other algorithms orders of magnitude slower. 
%
These properties make it ideal for real-time applications like Visual Odometry, SLAM, or self-localization in resource-limited devices.

%ELSED is an smart analysis of the image gradient, That is interpreted here as a map of the possibility that a certain region contains a segment, but this it can be replaced by a more elaborate map created using successive convolutions as is usual in CNNs.

\section*{Acknowledgements}

%The authors thank the anonymous reviewers for their comments.
This work was supported by Doctorado Industrial grant DI-16-08966 and MINECO project TIN2016-75982-C2-2-R.

\bibliographystyle{elsarticle-num} 
\bibliography{main.bib}

\end{document}